%% file: main.tex
\documentclass{CVM}
\usepackage{tabularx}
\usepackage{tikz}
\usepackage{xcolor} 
\usepackage{graphicx} 
\usetikzlibrary{trees,positioning,shapes,shadows,arrows, calc}
\usepackage[edges]{forest}
\usepackage{lmodern}
\usepackage{caption}
\usepackage{chronosys}

\catcode`\@=11
\def\chron@selectmonth#1{\ifcase#1\or January\or February\or
March\or April\or May\or June\or
July\or August\or September\or
October\or November\or December\fi}

\definecolor{hiddendraw}{RGB}{205, 44, 36}
\newcommand{\blu}[1]{\textcolor{black}{#1}}
\newcommand{\V}[1]{\mathbf{#1}}

\CVMsetup{
type      = {Review Article},
doi       = {CVM.XXXX},
title     = {NeRF: Neural Radiance Field in 3D Vision, A Comprehensive Review},
author    = {Kyle Gao$^{1}$, Yina Gao$^{2}$, Hongjie He$^{3}$, Dening Lu$^{1}$, Linlin Xu$^{4}$, and Jonathan Li$^{1,3}$ \cor{}\\},
runauthor = {K. Gao, et al.},
abstract  = {In March 2020, Neural Radiance Field (NeRF) revolutionized Computer Vision, allowing for implicit, neural network-based scene representation and novel view synthesis. NeRF models have found diverse applications in robotics, urban mapping, autonomous navigation, virtual reality/augmented reality, and more. In August 2023, Gaussian Splatting, a direct competitor to the NeRF-based volume rendering frameworks, was proposed. Gaussian Splatting gained tremendous momentum, overtaking NeRF-based methods as the dominant framework for novel view synthesis. We present a comprehensive survey of NeRF papers from the past five years (2020-2025). These include papers from the pre-Gaussian Splatting era, where NeRF and neural field rendering dominated the field of novel view synthesis and applications thereof. We also include works from the post-Gaussian Splatting era, where NeRF and implicit/hybrid neural fields found more niche applications. 

Our survey is organized into architecture and application-based taxonomies in the pre-Gaussian Splatting era, as well as a categorization of active research areas for NeRF, neural field, and implicit/hybrid neural representation methods. In the post-Gaussian Splatting era, we focus on relevant developments and applications. We provide an introduction to the theory of NeRF and its training via differentiable volume rendering. We also present a benchmark comparison of the performance and speed of classical NeRF, implicit and hybrid neural representation, and neural field models, and an overview of key datasets.},
keywords  = {Neural Radiance Field, NeRF, Neural Fields, Novel View Synthesis, Volume Rendering},
copyright = {The Author(s)},
}

\begin{document}

\maketitle

\enlargethispage{-3pt}
\begin{figure}[b] \vskip -4mm
\small\renewcommand\arraystretch{1.3}
\begin{tabular}{p{80.5mm}} \toprule\\ \end{tabular}
\vskip -4.5mm \noindent \setlength{\tabcolsep}{1pt}
\begin{tabular}{p{3.5mm}p{80mm}}
$1\quad $ & Department of Systems Design Engineering, University of Waterloo, Waterloo, ON N2L 3G1, Canada. E-mail: K. Gao, y56gao@uwaterloo.ca; D. Lu, d62lu@uwaterloo.ca; J. Li, junli@uwaterloo.ca \cor{}.\\
$2\quad $ & Faculty of Engineering, University of Toronto, Toronto, ON  M5S 1A1, Canada. E-mail: yina.gao@mail.utoronto.ca.\\
$3\quad $ & Department of Systems Design Engineering, University of Waterloo, Waterloo, ON N2L 3G1, Canada. E-mail: H. He, hongjie.he@uwaterloo.ca; J. Li, junli@uwaterloo.ca \cor{}.\\
$4\quad $ & Department of Geomatics Engineering, University of Calgary, Calgary, AB T2N 1N4, Canada. E-mail: lincoln.xu@ucalgary.ca.\\
&\hspace{-5mm} Manuscript received: 2025-X-X; accepted: 2025-X-X\vspace{-2mm}
\end{tabular} \vspace {-3mm}
\end{figure}

\input{1Intro}

\input{2Background}
\input{3NERFs}

\input{3_5NeRF_Applications}

\input{4PostGS}
\input{5.Discussion_and_conclusion}
\subsection*{Availability of Data and Material}
This review draws on published literature and public knowledge. All sources cited in the review are accessible through standard scholarly channels.

\subsection*{Acknowledgements and Funding}
This work was supported in part by the Natural Sciences and Engineering Research Council of Canada (NSERC) Discovery Grant (No. RGPIN-2022-03741).

\subsection*{Declaration of Competing Interest}

The authors have no competing interests to declare that are relevant to the
content of this article.\\

\subsection*{Author Contributions}
\textbf{Kyle Gao}  
Conceptualization. Literature review. Writing original draft. Writing review and editing.
\textbf{Yina Gao}  
Discussion. Writing review and editing.
\textbf{Hongjie He}  
Discussion. Writing review and editing.
\textbf{Dening Lu}  
Discussion. Writing review and editing.
\textbf{Linlin Xu}  
Supervision. Funding acquisition.
\textbf{Jonathan Li}  
Supervision. Funding acquisition.

\bibliographystyle{CVMbib}
\bibliography{bibfile}
\end{document}

%% file: 1Intro.tex
\section{Introduction}
Neural rendering leverages developments in deep learning for video generation and image synthesis. Implicit neural rendering methods use ``hidden" 3D representations (e.g., stored in a neural network as a neural field). Explicit neural rendering methods store 3D representations in explicit data structures. There are also hybrid methods that utilize both implicit and explicit 3D representations. Neural Radiance Fields (NeRF) use differentiable volume rendering to learn a (typically) implicit neural scene representation, using Multi-Layer Perceptrons (MLPs) to store the geometry and lighting of a 3D scene as neural fields. This learned representation can then be used to generate 2D images of the scene under novel, user-specified viewpoints (novel view synthesis). Mildenhall et al. first introduced NeRF at ECCV 2020 \cite{nerf2020_mildenhall}, and since then, it has achieved state-of-the-art visual quality, produced impressive demonstrations, and inspired many subsequent works. Since 2020, NeRF models and subsequent neural field-based volume rendering models have found applications in photo editing, 3D surface extraction, human avatar modeling, large/city-scale 3D representation and view synthesis, and 3D object generation.

In 2023, Gaussian Splatting \cite{2023gaussian_splatting}, an alternative novel view synthesis framework, overtook NeRF and adjacent methods on many novel view synthesis benchmarks, as well as in applications in 3D vision. As such, much of the research interest has shifted toward Gaussian Splatting. Nonetheless, research in NeRF and NeRF-adjacent neural rendering has persisted since 2023.

NeRF models have important advantages over classical methods (pre-NeRF and pre-Gaussian splatting) of novel view synthesis and scene representation.
\begin{itemize}
\item NeRF models are self-supervised. They can be trained using only multi-view images of a scene. Unlike many other neural representations of 3D scenes, NeRF models require only images and poses to learn a scene and do not require 3D or depth supervision. The poses can be estimated using Structure from Motion (SfM) packages such as COLMAP \cite{2016Colmap}.
\item NeRF models are photorealistic. Compared to classical techniques such as \cite{1996lfr} \cite{1996lumigraph}, as well as earlier novel view synthesis methods such as \cite{2019llf_forwardfacingdataset}\cite{2019srn}\cite{2019neuralvolumes}, and neural 3D representation methods \cite{2020differentiableneuralrendering}\cite{2020local3D_occupency}\cite{2019deep_sdf}, the original NeRF model converged to better results in terms of visual quality, with more recent models performing even better.
\end{itemize}
\begin{figure*}[h!] 
\centering
\includegraphics[width=0.9\textwidth]{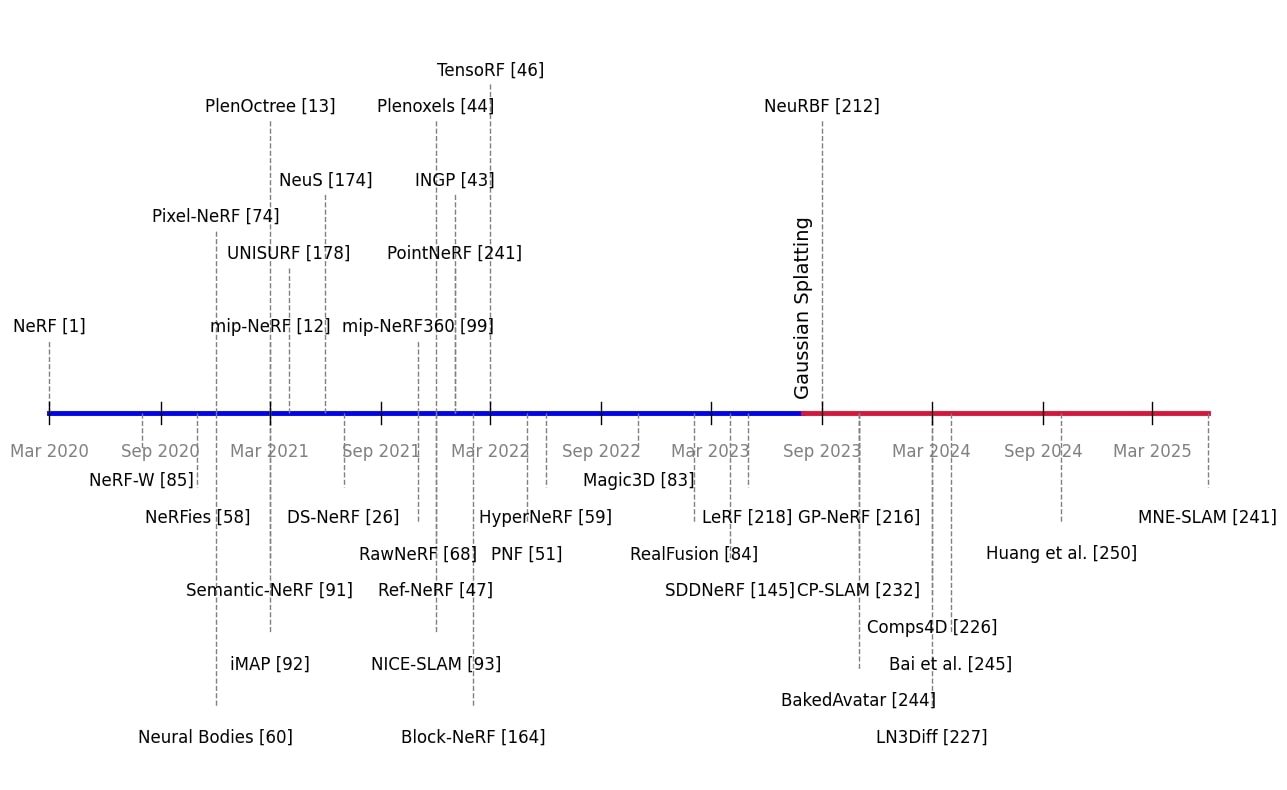}
\caption{Timeline of important and influential NeRF and neural volume rendering methods pre- and post-Gaussian Splatting. Above the line are general or fundamental methods used by future subsequent methods to build neural fields. Below the line are various application-specific methods.} \label{fig:timeline}
\end{figure*}
Compared to Gaussian Splatting-based methods, which have largely overtaken NeRF-based methods in novel view synthesis and adjacent research in 2023-2025, NeRF methods have the following disadvantages.
\begin{itemize}
\item Gaussian Splatting methods are more photorealistic than NeRF methods, typically converging to representations that generate higher-quality images.
\item Gaussian Splatting methods are faster to train. On equal hardware, using the same training images, fully implicit NeRF methods take two to three orders of magnitude longer to converge. Once fully trained, Gaussian Splatting methods also render images faster than implicit NeRF-based methods by orders of magnitude.
\item Gaussian Splatting methods use 3D point-based representations and can be easily converted into 3D point clouds, a common data structure for representing 3D scenes. On the other hand, extracting an explicit 3D representation from a typical NeRF method is more difficult.
\end{itemize}
However, NeRF methods have the following advantages over Gaussian Splatting methods.
\begin{itemize}
\item An implicit or hybrid NeRF method has lower storage requirements after training and usually lower memory requirements during training.
\item NeRF methods, with neural network-based 3D representations, are better suited for 3D vision pipelines that require or prefer implicit representations.
\end{itemize}

We organized a survey paper focusing on NeRF methods and NeRF-like neural rendering methods, extending our earlier highly cited preprint on works from 2020 to 2023 with post-Gaussian Splatting (post-GS) methods from 2023 to 2025. A timeline of important methods is shown in Fig. \ref{fig:timeline}.

In the post-Gaussian Splatting era, many implicit and hybrid neural representation methods distance themselves from the keyword “NeRF” and instead emphasize the terms neural field and neural representation. We include these methods in the post-Gaussian Splatting chapter if they use NeRF-like differentiable volume rendering in conjunction with a neural field-based representation. We do not make an explicit distinction pre-Gaussian Splatting, since many of these adjacent methods' neural field-based volume rendering techniques were referred to as NeRF or NeRF-like pre-Gaussian Splatting, even if they used other types of neural fields such as neural signed distance functions or occupancy fields. Post-Gaussian Splatting, we use the same naming conventions as the authors of the respective papers.

The rest of this manuscript is organized as follows.
\begin{itemize}
\item Section \ref{section:background} introduces existing NeRF survey preprints (\ref{subsection:survey}), explains the theory behind NeRF volume rendering (\ref{subsection:nerf_theory}), and introduces commonly used datasets (\ref{subsection:dataset}) and quality assessment metrics (\ref{subsection:QA_metrics}).
\item Sections \ref{sec:nerf} and \ref{Applications} form the first part of the main body of the survey. Section \ref{sec:nerf} introduces influential NeRF publications from before the Gaussian Splatting era and presents the taxonomy we created to organize these works. Its subsections detail the different families of NeRF and adjacent methods’ innovations proposed from 2020 to 2023.
\item Section \ref{Applications} focuses on the applications of NeRF and adjacent models to various computer vision tasks.
\item Section \ref{sec:postgs} forms the second part of the main body of the survey. This section introduces progress in neural rendering research post-Gaussian Splatting (2023 onward). We specifically focus on post-Gaussian Splatting applications of implicit and hybrid neural field methods that use NeRF-like volume rendering.
\item Sections \ref{sec:discussion} and \ref{sec:conclusion} present discussions on the status of this research field post-Gaussian Splatting and provide concluding remarks.
\end{itemize}

%% file: 2Background.tex
\section{Background} \label{section:background}
\subsection{Existing Surveys on NeRF with Comparable Scope} \label{subsection:survey}

In December 2020, Dellaert and Yen-Chen published a concise preprint NeRF survey \cite{dellaert2020nerfSurvey} including approximately 50 NeRF publications/preprints, many of which were eventually published in top-tier computer vision conferences. We took inspiration from this preprint survey and used it as a starting point for our own survey. However, the work is only eight pages long and does not include detailed descriptions. Moreover, it only includes NeRF papers from 2020 and early 2021 preprints and omits several influential NeRF papers published in the latter half of 2021 and beyond.

In November 2021, Xie et al. \cite{xie2022survey} published a preprint survey titled Neural Fields in Visual Computing and Beyond, which was later revised and published as a state-of-the-art report in Eurographics 2022, and also presented as a tutorial at CVPR 2022. The survey is broad and includes a detailed technical introduction to the use of neural fields in computer vision, as well as applications in visual computing, with a strong focus on Neural Radiance Fields. Compared to this work, we restrict our review to NeRF papers only and include more recent work. We also present more detail on a paper-by-paper basis.

In December 2021, Zhang et al. \cite{2021multimodal_survey} published a preprint survey on multimodal image synthesis and editing, in which they dedicated a paragraph to NeRF. They mostly focused on multimodal NeRFs such as \cite{2022clipnerf}, \cite{2021adnerf}, only citing these two and the original NeRF paper \cite{nerf2020_mildenhall} in the main text, with four additional papers \cite{2022zerotextnerf_surv} \cite{2021cgnerf_surv} \cite{2022idenerf_surv} \cite{2022sem2nerf_surv} included in their supplementary materials.

In May 2022, Tewari et al. \cite{2022advances_survey} published a state-of-the-art report on advances in neural rendering with a focus on NeRF models. It remains the most comprehensive neural rendering survey-style report to date, covering many influential NeRF papers, as well as numerous other neural rendering papers. Our survey differs from this report in that our scope is fully focused on NeRF papers, providing detailed paper-by-paper summaries of selected works. We also present a NeRF innovation technique taxonomy tree and a NeRF application classification tree. We include more recent works and introduce, in detail, the common datasets and evaluation metrics used by NeRF practitioners.

These survey papers were concurrent with our initial manuscript, completed in 2022, and are most similar in scope. Since the Gaussian Splatting era, numerous surveys focusing on specific applications of neural fields have been completed. However, these surveys differ from ours in scope.

\subsection{Neural Radiance Field (NeRF) Theory} \label{subsection:nerf_theory}


Neural Radiance Fields were first proposed by Mildenhall et al. \cite{nerf2020_mildenhall} in 2020  for novel view synthesis. NeRFs achieved highly photo-realistic view synthesis of complex scenes and attracted much attention in the field. In its basic form, a NeRF model represents three-dimensional scenes as a radiance field approximated by a neural network. The radiance field describes color and volume density for every point and for every viewing direction in the scene. This is written as:

\begin{equation} \label{eq:nerf}
    F(\mathbf{x},\theta,\phi) \xrightarrow{} (\mathbf{c},\sigma),
\end{equation}
where $\mathbf{x}=(x,y,z)$ is the in-scene coordinate, $(\theta,\phi)$ represent the azimuthal and polar viewing angles, $\mathbf{c} = (r,g,b)$ represents color, and $\sigma$ represents the volume density. This 5D function is approximated by one or more Multi-Layer Perceptrons (MLPs), sometimes denoted as $F_\Theta$. The two viewing angles $(\theta,\phi)$ are often represented by $\mathbf{d}=(d_x,d_y,d_z)$, a 3D Cartesian unit vector. This neural network representation is constrained to be multi-view consistent by restricting the prediction of $\sigma$, the volume density (i.e., the content of the scene), to be independent of viewing direction, whereas the color $\mathbf{c}$ is allowed to depend on both viewing direction and in-scene coordinate. In the baseline NeRF model,  this is implemented by designing the MLP to be in two stages. The first stage takes as input $\mathbf{x}$ and outputs $\sigma$ and a high-dimensional feature vector (256 in the original paper). In the second stage, the feature vector is then concatenated with the viewing direction $\mathbf{d}$, and passed to an additional MLP, which outputs $\mathbf{c}$. We note that Mildenhall et al. \cite{nerf2020_mildenhall} consider the $\sigma$ MLP and the $\V{c}$ MLP to be two branches of the same neural network, but many subsequent authors consider them to be two separate MLP networks. 

Broadly speaking, novel view synthesis using a trained NeRF model is as follows.
\begin{itemize}
    \item For each pixel in the image being synthesized, send camera rays through the scene and generate a set of sampling points.
    \item For each sampling point, use the viewing direction and sampling location to compute local color and density using the NeRF MLP(s).
    \item Use volume rendering to produce the image from these colors and densities.
\end{itemize}

Given the volume density and color functions of the scene being rendered, volume rendering \cite{NERF1989_Raytracing} is used to obtain the color $C(\mathbf{r})$ of any camera ray $\mathbf{r}(t) = \mathbf{o} + t\mathbf{d}$, with camera position $\mathbf{o}$ and viewing direction $\mathbf{d}$ using

\begin{equation} \label{eq:ray}
    C(\mathbf{r}) = \int^{t_2}_{t_1} T(t) \cdot \sigma(\mathbf{r}(t)) \cdot \mathbf{c}(\mathbf{r}(t),\mathbf{d}) \cdot dt,
\end{equation}
where $\sigma(\mathbf{r}(t))$ and $\mathbf{c}(\mathbf{r}(t),\mathbf{d})$ represent the volume density and color at point $\mathbf{r}(t)$ along the camera ray with viewing direction $\mathbf{d}$, and $dt$ represents the differential distance traveled by the ray at each integration step. \blu{The integration bounds $(t_1,t_2)$ are defined such that the near bound $t_1$ parametrizes the camera center $\mathbf{r}(t_1)$, and the far bound $t_2$ is determined by the scene's bounding volume.}  

$T(t)$ is the accumulated transmittance, representing the probability that the ray travels from $t_1$ to $t$ without being intercepted, given by

\begin{equation} \label{eq:transmisivity}
    T(t) = \exp(-\int^{t}_{t_1}  \sigma(\mathbf{r}(u))\cdot du).
\end{equation}

Novel views are rendered by tracing the camera rays $C(\mathbf{r})$ through each pixel of the to-be-synthesized image. This integral can be computed numerically. The original implementation \cite{nerf2020_mildenhall} and most subsequent methods used a non-deterministic stratified sampling approach, where the ray was divided into $N$ equally spaced bins, and a sample was uniformly drawn from each bin. Then, equation (\ref{eq:ray}) can be approximated as

\begin{equation} \label{eq:ray_discrete}
    \hat{C}(\V{r}) = \sum_{i=1}^N \alpha_i T_i \V{c}_i, \; \text{where} \; T_i = \exp(-\sum_{j=1}^{i-1} \sigma_j \delta_j).
\end{equation}
$\delta_i$ is the distance from sample $i$ to sample $i+1$. $(\sigma_i, \V{c_i})$ are the density and color evaluated along the sample point $i$ given the ray, as computed by the NeRF MLP(s). $\alpha_i$, the transparency/opacity from alpha compositing at sample point $i$, is given by 

\begin{equation} \label{eq:alpha}
    \alpha_i = 1-\exp(-\sigma_i \delta_i).
\end{equation}
For each pixel, a squared error photometric loss is used to optimize the MLP parameters. Over the entire image, this is given by

\begin{equation} \label{eq:photo-loss}
   L = \sum_{r \in R} || \hat{C}(\V{r}) - C_{gt}(\V{r})||_2^2,
\end{equation}
where $C_{gt}(\V{r})$ is the ground truth color of the training image's pixel associated with $\V{r}$, and $R$ is the batch of rays associated with the to-be-synthesized image.
An expected depth can be calculated for the ray using the accumulated transmittance \blu{by integrating near bound $t_1$ at the camera center, to the far bound $t_2$ at the scene’s boundary:}
\begin{equation} \label{eq:expected depth}
    d(\V{r}) = \int^{t_2}_{t_1} T(t) \cdot \sigma(\mathbf{r}(t)) \cdot t \cdot dt.
\end{equation}
This can be approximated analogously to how Eq. (\ref{eq:ray_discrete}) approximates Eq. (\ref{eq:ray}) and Eq. (\ref{eq:transmisivity})
\begin{equation}
     \hat{D}(\V{r}) = \sum_{i=1}^N \alpha_i t_i T_i.
\end{equation}

 Certain depth regularization \cite{2022regnerf} \cite{2022dsnerf} \cite{2022nerfren} \cite{2022sinnerf} methods use the expected depth to restrict densities to delta-like functions at scene surfaces, or to enforce depth smoothness. The additional depth supervision term also helps the geometry converge earlier and with fewer input views.
 
NeRF models often employ positional encoding, which was shown by Mildenhall et al. \cite{nerf2020_mildenhall} to greatly improve fine detail reconstruction in the rendered views. This was also shown in more detail, with corroborating theory using Neural Tangent Kernels in \cite{2020PosEnc}.  In the original implementation, the following positional encoding $\gamma$ was applied to each component of the scene coordinate  $\mathbf{x}$ (normalized to [-1,1]) and viewing direction unit vector $\mathbf{d}$
\begin{multline}
    \gamma(v) = (sin(2^0 \pi v), cos(2^0 \pi v), sin(2^1 \pi v),  cos(2^1 \pi v), \\
    ..., sin(2^{N-1} \pi v), cos(2^{N-1} \pi v)),
\end{multline}
where N is a user-determined encoding dimensionality parameter, set to $N=10$ for $\mathbf{x}$ and $N=4$ for $\mathbf{d}$ in the original paper. However, modern researchers have experimented and achieved great results with alternate forms of positional encoding including trainable parametric, integral, and hierarchical variants (see section \ref{sec:nerf}).

Naming conventions: There are three types of 3D representation: implicit, hybrid, and explicit. In the baseline NeRF, the density and color fields are fully represented by MLPs; this is considered an implicit scene representation. The neural color and density fields together are called the neural radiance field. Methods with hybrid and explicit scene representations are introduced in Sections \ref{sec:speed} and \ref{sec:no mlp}, respectively. Other types of neural fields, such as neural occupancy fields, neural signed distance function fields, neural deformation fields, and neural semantic fields, are introduced throughout this survey.

\subsection{Datasets} \label{subsection:dataset}

NeRF models are typically trained per-scene and require relatively dense images with relatively varied poses. While some NeRF models have been designed to be trained from sparse input views or unposed images, camera poses can often be extracted using existing Structure-from-Motion libraries such as COLMAP \cite{2016Colmap}.

The original NeRF paper \cite{nerf2020_mildenhall} presented a synthetic dataset created using Blender (referred to as the \textbf{Realistic Synthetic 360 Degrees dataset} in \cite{nerf2020_mildenhall}), and often referred to as the \textbf{NeRF Synthetic dataset or the NeRF dataset} in subsequent works. The virtual cameras have the same focal length and are placed at the same distance from the object. The dataset is composed of eight scenes with eight different objects. For six of these, viewpoints are sampled from the upper hemisphere; for the other two, viewpoints are sampled from the entire sphere. The objects are ``hotdog", ``materials", ``ficus", ``lego", ``mic", ``drums", ``chair", and ``ship". The images are rendered at 800$\times$800 pixels, with 100 views for training and 200 views for testing. This is often the first dataset considered by NeRF researchers, as the scenes are bounded, focused on a single object, and benchmarks of commonly used models on this scene are easily found.

The \textbf{Mip-NeRF 360 dataset} \cite{2022mip360} comprises nine real-world scenes, including five outdoor (``bicycle", ``flowers", ``garden", ``stump", ``treehill") and four indoor (``bonsai", ``counter", ``kitchen", ``room") environments. Each scene contains between 120 and 300 high-resolution, camera-posed images captured along a complete 360-degree trajectory encircling the main region of interest. This setup enables full coverage of both the central object and its extended surroundings. Unlike the original NeRF dataset, which focuses on bounded and object-centric captures, the Mip-NeRF 360 dataset introduces complex, unbounded backgrounds with significant depth variation and lighting diversity. Consequently, it serves as a challenging benchmark for evaluating neural rendering and novel-view synthesis methods capable of handling large-scale, unbounded 3D scenes.

The \textbf{LLFF dataset} \cite{2019llf_forwardfacingdataset} consists of 24 real-life scenes captured from handheld cellphone cameras. The views are forward-facing towards the central object. Each scene consists of 20-30 images. The COLMAP \cite{2016Colmap} package was used to compute the poses of the images. The usage of this dataset is comparable to that of the Realistic Synthetic dataset from \cite{nerf2020_mildenhall}; the scenes are not too challenging for any particular NeRF model, and the dataset is well benchmarked, offering readily available comparisons to known methods.

The \textbf{DTU dataset} \cite{2014DTUdataset} is a multi-view stereo dataset captured using a 6-axis industrial robot mounted with both a camera and a structured light scanner. The robot provides precise camera positioning. Both the camera intrinsics and poses are carefully calibrated using the MATLAB calibration toolbox \cite{matlab_calibration}. The light scanner provides reference dense point clouds, which serve as ground truth 3D geometry. Nonetheless, due to self-occlusion, the scans of certain areas in some scenes are incomplete. The original paper's dataset consists of 80 scenes, each containing 49 views sampled on a sphere of radius 50 cm around the central object. For 21 of these scenes, an additional 15 camera positions are sampled at a radius of 65 cm, for a total of 64 views. The entire dataset includes 44 additional scenes that have been rotated and scanned four times at 90-degree intervals. The illumination of scenes is varied using 16 LEDs, with seven different lighting conditions. The image resolution is 1600 $\times$ 1200. This dataset differs from the previous ones by its higher resolution and carefully calibrated camera motion and poses.

The \textbf{ScanNet dataset} \cite{2017scannet} is a large-scale real-life RGB-D multi-modal dataset containing more than 2.5 million views of indoor scenes, with annotated camera poses, reference 3D surfaces, semantic labels, and CAD models. The depth frames are captured at 640 $\times$ 480 pixels, and the RGB images are captured at 1296 $\times$ 968 pixels. The scans were performed using RGB-D sensors attached to handheld devices such as iPhones and iPads. The poses were estimated using BundleFusion \cite{2017bundlefusion} and geometric alignment of the resulting mesh. This dataset’s rich semantic labels are useful for models that utilize semantic information, such as for scene editing, scene segmentation, and semantic view synthesis.

The \textbf{ShapeNet dataset} \cite{2015shapenet} is a simple large-scale synthetic 3D dataset consisting of 3D CAD models classified into 3,135 classes. The most commonly used subset is the 12 common object categories. This dataset is sometimes used when object-based semantic labels are an important part of a particular NeRF model. From ShapeNet CAD models, software such as Blender is often used to render training views with known poses.

\subsubsection{Building-scale Dataset}

The \textbf{Tanks and Temples dataset} \cite{2017tanksandtemples} is a from-video 3D reconstruction dataset. It consists of 14 scenes, including individual objects such as ``Tank" and ``Train", and large-scale indoor scenes such as ``Auditorium" and ``Museum." Ground truth 3D data was captured using a high-quality industrial laser scanner. The ground truth point cloud was used to estimate camera poses using least squares optimization of correspondence points. This dataset contains large-scale scenes, some of which are outdoors, and poses a challenge for certain NeRF models. The outdoor scenes are suited for models wishing to handle unbounded backgrounds. Its ground truth point cloud can also be used for certain data fusion methods or to test depth reconstruction.

\textbf{Matterport-3D dataset} \cite{2017matterport3d} is a real-life dataset consisting of 10,800 panoramic views from 194,400 RGB-D globally registered images of 90 building-scale scenes. Depth, semantic, and instance annotations are available. Color and depth images are provided at 1280 $\times$ 1024 resolution for 18 viewpoints per panoramic picture. Each of the 90 buildings consists of an average of 2,437 m$^2$ of surface area. A total of 50,811 object instance labels were provided, which were mapped into 40 object categories.

The \textbf{Replica dataset} \cite{2019replica} is a real indoor dataset consisting of 18 scenes and 35 indoor rooms captured using a custom-built RGB-D rig with an IR projector. Certain 3D features were manually fixed, such as fine-scale mesh details like small holes, and reflective surfaces were manually assigned. Semantic annotations (88 classes) were performed in two steps: once in 2D and once in 3D. Both class-based and instance-based semantic labels are available.

The \textbf{Deep Blending dataset} \cite{2018deepbleding} consists of a collection of 19 real-world indoor and outdoor scenes designed for free-viewpoint image-based rendering (IBR). It provides two complementary data partitions: (1) \textit{Reconstruction inputs and outputs}, which include the raw input images, COLMAP-based structure-from-motion (SfM) and multi-view stereo (MVS) reconstructions, a global textured mesh and refined depth maps; and (2) \textit{IBR inputs and outputs}, which contain the global textured mesh, reduced SfM reconstruction, per-view meshes, input camera poses in the NVM format, and test camera trajectories with corresponding rendered results. Compared to object-centric NeRF-style datasets, the Deep Blending dataset provides richer geometric context, realistic lighting variation, and full reconstruction pipelines, establishing a benchmark for free-viewpoint image-based rendering in structured, real-world environments.

\input{figures/table1}
\subsubsection{Large-Scale Urban Datasets}
Popular autonomous driving benchmark datasets have multiple data modalities such as images, depth maps, LiDAR point clouds, poses, and semantic maps, which are potentially suitable for certain NeRF models targeting urban scenes. Recently, NeRF models such as those proposed in \cite{2021mine} and \cite{2022panoptic} have made effective use of these datasets.

\textbf{KITTI} \cite{2012kitti}\cite{2013kittidataset}\cite{kitti013ITSC}\cite{kitti2015CVPR} is a well-known city-scale 2D-3D computer vision dataset suite created for training and benchmarking vision algorithms for autonomous driving. The suite contains labeled datasets for stereo 3D semantic and 2D semantic segmentation, flow, odometry, 2D-3D object detection, tracking, lane detection, and depth prediction/completion. These were created from raw LiDAR and video data captured in Karlsruhe, Germany, using a car-based setup with GPS and inertial measurement unit data, recorded with a Velodyne LiDAR scanner and multiple cameras. The depth prediction/completion dataset is by far the largest, containing over 93 thousand depth maps with corresponding RGB images and raw LiDAR scans. This dataset, however, poses a challenge to NeRF training due to its relatively sparse camera coverage compared to NeRF-specific datasets, requiring sparse-view considerations when designing the model. The recent KITTI-360 \cite{liao2022kitti} extension to the suite includes a novel view synthesis benchmark which tabulates many NeRF models.

The \textbf{Waymo open dataset} \cite{2020waymo} is a recently published alternative to KITTI. Covering 72km$^2$, the dataset is created from point cloud and video data captured from five LiDAR sensors and five high-resolution pinhole cameras in a car-based setup, captured in the San Francisco Bay, Mountain View, and Phoenix in the United States. In addition to matched point cloud and video data, the dataset also contains annotated labels for 2D and 3D object detection and tracking. The dataset contains 1150 separate scenes (as opposed to KITTI's 22 scenes) and has higher LiDAR and camera resolution. Its object annotations are also more extensive by two orders of magnitude (80K vs 12M). 

The \textbf{BungeeNeRF dataset} \cite{2022bungeenerf} contains large-scale outdoor scenes that capture urban environments across a wide range of spatial scales. The data are sourced from Google Earth Studio and each scene is recorded using concentric circular camera trajectories centered around a key landmark or building of interest. These trajectories vary in radius and elevation, providing multi-scale coverage from distant aerial views to close ground-level perspectives. Each scene includes high-resolution images and precise camera poses to support the study of neural radiance fields under extreme scale variation.

\subsubsection{Human Avatar/Face Dataset}
The \textbf{Nerfies} \cite{nerf2021_nerfies} and \textbf{HyperNerf} \cite{nerf2021_hypernerf} datasets are single-camera datasets focused on human faces, with motion generated by moving two cameras attached to a pole relative to the subject. The former contains five human subjects staying still, as well as four more scenes with moving human subjects, a dog, and two moving objects. The latter focuses on topological changes and includes scenes such as a human subject opening and closing their eyes and mouth, peeling a banana, 3D printing a chicken toy, and a broom deforming.

The \textbf{ZJU-MOCap LightStage dataset} \cite{2021neuralbody} is a multi-view (20+ cameras) motion capture dataset consisting of 9 dynamic human sequences consisting of exercise-like motions. The videos were captured using 21 synchronized cameras and have a sequence length between 60 to 300 frames. 

The \textbf{NeuMan dataset} \cite{2022neuman} consists of 6 videos, each 10 to 20 seconds long, captured by a mobile phone camera following a walking human subject performing additional simple actions such as twirling or waving.

The \textbf{CMU Panoptic dataset} \cite{2015panoptic} is a large multi-view, multi-subject dataset consisting of groups of people engaged in social interaction. The dataset contains 65 sequences with 1.5 million labeled skeletons. The sensor system consists of 480 VGA views (640×480), over 30 HD (1920×1080) views, and 10 RGB-D sensors. Scenes are labeled with individual subject and social group semantics, 3D body poses, 3D facial landmarks, and transcripts with speaker IDs.

\subsection{Quality Assessment Metrics}\label{subsection:QA_metrics}
Novel view synthesis via NeRF uses visual quality assessment metrics for benchmarks. These metrics attempt to assess the quality of individual images either with (full-reference) or without (no-reference) ground truth images. Peak Signal to Noise Ratio (PSNR), Structural Similarity Index Measure (SSIM) \cite{2004ssim}, Learned Perceptual Image Patch Similarity (LPIPS) \cite{2018lpips} are by far the most commonly used in NeRF literature. 

PSNR$\uparrow$ is a full reference quality assessment metric,
\begin{equation}
    PSNR(I) = 10 \cdot \log_{10}(\frac{MAX(I)^2}{MSE(I)}),
\end{equation}
where $MAX(I)$ is the maximum possible pixel value in the image (255 for 8bit integer), and $MSE(I)$ is the pixel-wise mean squared error calculated over all color channels. PSNR is also commonly used in signal processing and is well understood.

SSIM$\uparrow$ \cite{2004ssim} is a full-reference quality assessment metric. For a single patch, this is given by
\begin{equation}
    SSIM(x,y) = \frac{(2\mu_x \mu_y + C_1)(2\sigma_{xy} + C_2)} {(\mu_x^2+\mu_y^2 + C_1)(\sigma_x^2+\sigma_y^2 + C_2)},
\end{equation}
where $C_i = (K_iL)^2$, $L$ is the dynamic range of the pixels (255 for 8bit integer), and $K_1 = 0.01, K_2 = 0.03$ are constants chosen by the original authors. We note that there is a more general form of SSIM given by (12) in the original paper. 
The local statistics $\mu's, \sigma's$ are calculated within an $11 \times 11$ circular symmetric Gaussian weighted window, with weights $w_i$ having a standard deviation of 1.5 and normalized to 1. 

LPIPS$\downarrow$ \cite{2018lpips} is a full reference quality assessment metric which uses learned convolutional features. The score is given by a weighted pixel-wise MSE of feature maps over multiple layers.
\begin{equation}
    LPIPS(x,y) = \sum_l^L \frac{1}{H_l W_l} \sum_{h,w}^{H_l, W_l} ||w_l \odot (x_{hw}^l - y_{hw}^l)||^2_2,
\end{equation}
where $x_{hw}^l,y_{hw}^l$ are the reference and assessed images' feature at pixel width $w$, pixel height $h$, and layer $l$. $H_l$ and $W_l$ are the feature maps height and width at the corresponding layer. The original LPIPS paper used SqueezeNet \cite{2016squeezenet}, VGG \cite{2014vgg} and AlexNet \cite{2012alexnet} as feature extraction backbone. 

%% file: figures/table1.tex
\begin{table*}[t]
\centering
\caption{Comparison of select NeRF models on the Synthetic NeRF dataset
\cite{nerf2020_mildenhall}} \label{tab:benchmark}
\scriptsize
\begin{tabularx}{\textwidth}{l|ccccccc}
Method        & Representation & PSNR$\uparrow$ (dB) & SSIM$\uparrow$ & LPIPS$\downarrow$ & Training Iteration & Training Time$^{1}$ & Inference Speed$^{2}$ \\ \hline
Baseline NeRF (2020) \cite{nerf2020_mildenhall}      & Implicit         & 31.01    & 0.947    & 0.081     &100-300k              & $>$12h  & 1        \\ \hline
\multicolumn{8}{c}{Speed Improvement}
\\ \hline
JaxNeRF (2020) \cite{jaxnerf2020github}                    & Implicit           & 31.65    & 0.952    & 0.051     &250k              & $>$12h  & $\sim$1.3        \\ \hline
NSVF (2020) \cite{nerf2020_NSVF}                   & Hybrid (Learned)             & 31.74    & 0.953    & 0.047     &  100-150k            & - &$\sim$10                  \\ \hline

SNeRG (2021) \cite{2021sNeRG}                 & Explicit (Baked)                & 30.38    & 0.950    & 0.050     &  250k            &$>$12h &$\sim$9000                  \\ \hline 

PlenOctree (2021) \cite{nerf2021_plenoctrees}     & Explicit (Baked)       & 31.71    & 0.958    & 0.053     &  2000k            & $>$12h &   $\sim$3000  \\ \hline

FastNeRF (2021) \cite{2021fastNerf}    & Explicit (Baked)         & 29.97    & 0.941    & 0.053     &  300k            & $>$12h &   $\sim$4000  \\ \hline

KiloNeRF (2021) \cite{nerf2021_kilonerf}   & Explicit (Baked)        & 31.00   & 0.95    & 0.03     &  600k+150k+1000k  &$>$12h &$\sim$2000      \\ \hline 

Instant-NGP (2022) \cite{nerf2022_ngp}   &  Hybrid (Learned)       & 33.18    & -    & -     &  256k            & $\sim$5m & "orders of magnitude"  \\ \hline

Plenoxels (2021) \cite{2021plenoxels}   &  Explicit (Learned)       & 31.71     & 0.958    &  0.049     &  128k    & $\sim$12m & -  \\ \hline

DVGO (2021) \cite{2022voxeldirect}   &  Explicit (Learned)       & 31.95-32.80     & 0.957-0.961    &  0.053-0.27     &  128k    & $\sim$20m & ~45x  \\ \hline

TensoRF (2022) \cite{2022tensorf}  & Explicit/Hybrid (Learned)& 31.56-33.14 & 0.949-0.963 & - & 15K-30K & ~8-25m & $\sim$100\\ \hline

\multicolumn{8}{c}{Quality Improvement}
\\ \hline

mip-NeRF (2021) \cite{2021mipnerf}     & Implicit       & 33.09    & 0.961    & 0.043     & 1000k              & $\sim$3h  &$\sim$1  \\ \hline 
ref-NeRF (2021) \cite{2022refnerf}      & Implicit             & 35.96    & 0.967    & 0.058     & 250k            &  - &$\sim$1  \\ \hline
\multicolumn{8}{c}{Sparse View/Few Shots}
\\ \hline

MVSNeRF (2021) \cite{2021mvsnerf}          & Hybrid + Pretrained & 27.07    & 0.931    & 0.163    & 10k (*3 views)            &  $\sim$15m* &$\sim$1  \\ \hline

DietNeRF (2021) \cite{2021dietNeRF}              & Implicit + Pretrained & 23.15    & 0.866    & 0.109    & 200k (*8 views)            &  - &$\sim$1  \\ \hline

DS-NeRF (2021) \cite{2022dsnerf}               & Implicit & 24.9    & 0.72   & 0.34    & 150-200k (*10 views)            &  - &$\sim$1  \\ \hline
\end{tabularx}
\flushleft
\small
Key speed, sparse-view and quality improvement models focused on small-scale scenes trained on the Synthetic NeRF dataset were selected for comparison.

\footnotesize
$^1$Training speeds are given as in the respective original papers. These are to be taken with "a grain of salt", as hardware differences and hyperparameters such as image/voxel resolution greatly affect training time. 
\footnotesize \\
$^2$Inference speeds are given as speedup factor over the baseline NeRF.
\\

\end{table*}

%% file: 3NERFs.tex

\input{figures/figure_innovation_taxonomy}

\section{Neural Radiance Field (NeRF) pre Gaussian Splatting} 
\label{sec:nerf}

In this section, we present select pre-Gaussian Splatting NeRF and adjacent papers organized in a method-based taxonomy, with a separate section reserved for application-based classification. The (arXiv) preprint first draft dates are used to sequence the publications, while conference/journal publication dates can be found in the bibliography. We included a NeRF synthetic dataset \cite{nerf2020_mildenhall} benchmark in Table \ref{tab:benchmark} comparing the most influential (pure) novel-view-synthesis-focused NeRF, Neural Rendering, and adjacent works of that era.

Subsection \ref{sec:fundamentals} studies models that sharpen photometric fidelity and refine learned geometry.
Subsection \ref{sec:speed} looks at schemes that improve training speed and inference speed through baked (distilling a NeRF model trained on a specific scene into a fast representation) or non-baked pipelines (e.g. changing/improving the NeRF framework in some fundamental way).
Subsection \ref{sec:sparse} details sparse or few-shot radiance fields that rely on learned priors and depth cues.
Subsection \ref{sec:conditional} follows generative and conditional fields that enable text-driven or image-driven editing through GAN, Diffusion, or global latent optimization methods.
Subsection \ref{sec:composition} examines scenes with compositional strategies that range from plain background separation to object-level semantic decomposition in rich environments and includes a subsection for 4D/dynamic scenes and methods for scene animation. 
Subsection \ref{sec:pose} reviews pose estimation that uses simple bundle adjustment or full SLAM-style systems.
Subsection \ref{sec:adjacent} covers adjacent differentiable volume rendering models built on explicit voxel fields or tensor fields and ray transformers.

\subsection{Improvements in the Quality of Synthesized Views and Learned Geometry} \label{sec:fundamentals}
Image quality is the predominant benchmark for view synthesis, and many subsequent pure NeRF research models have focused on improving view synthesis quality. In this section, we highlight important models that aim to enhance the photometric and geometric aspects of NeRF view synthesis and 3D scene representation.

\subsubsection{Better View Synthesis}

\textbf{Mip-NeRF} (March 2021) \cite{2021mipnerf} approximated cone tracing instead of using the ray tracing of standard NeRF \cite{nerf2020_mildenhall} (March 2020) volume rendering. This was achieved by introducing the Integrated Positional Encoding (IPE). Schematically, to generate an individual pixel, a cone was cast from the camera center along the viewing direction through the center of the pixel. The cone was approximated by a multivariate Gaussian whose mean vector and covariance matrix were derived to match the appropriate geometry (see Appendix A in \cite{2021mipnerf}), resulting in the Integrated Positional Encoding. This is given by

\begin{equation}
  \begin{array}{l}
    \gamma(\boldsymbol{\mu},\V{\Sigma}) =  \mathbb{E}_{\V{x} \sim \mathcal{N}}(\boldsymbol{\mu},\V{\Sigma}) [\gamma(\V{x})] 
    \\
    =\begin{bmatrix}
    \sin(\boldsymbol{\mu_\gamma})\odot \exp(-(1/2)\text{diag}(\boldsymbol{\Sigma_\gamma})) \\
    \cos(\boldsymbol{\mu_\gamma})\odot \exp(-(1/2)\text{diag}(\boldsymbol{\Sigma_\gamma})), 
    \end{bmatrix}
 \end{array}
\end{equation}
where $\boldsymbol{\mu_\gamma},\boldsymbol{\Sigma_\gamma}$ are the means and variances of the multivariate Gaussian lifted onto the positional encoding basis with $N$ levels. 
The resulting Mip-NeRF model was inherently multi-scale and performed anti-aliasing automatically. It outperformed the baseline NeRF \cite{nerf2020_mildenhall}, especially at lower resolutions. \textbf{Mip-NeRF 360} (November 2021) \cite{2022mip360} is a highly influential extension of Mip-NeRF to unbounded scenes. The key technical improvements include a proposal MLP supervised by the NeRF MLP rather than directly by images. This proposal MLP predicts only volumetric density, which guides the sampling intervals. Additionally, a novel scene parameterization was specifically designed for the Gaussians in Mip-NeRF. Finally, a new regularization method was introduced to prevent floating geometric artifacts and background collapse.

\begin{figure}[h!] 
\centering
\includegraphics[width=0.45\textwidth]{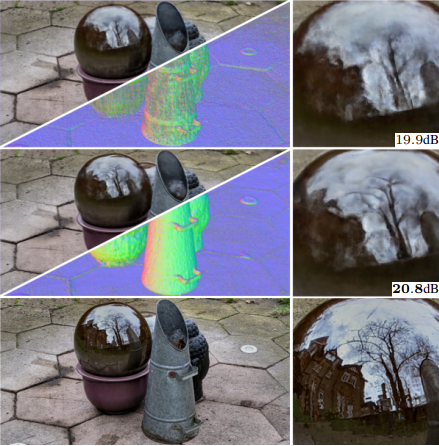}
\caption{Ref-NeRF results in the ``garden spheres" scene of \cite{2021sNeRG}, showing its high performance on reflective scenes and its capability to recover accurate normal vectors of reflective surfaces. Top: mip-NeRF \cite{2021mipnerf}; middle: ref-NeRF \cite{2022refnerf}; bottom: ground truth. ©2022 IEEE} \label{fig:refnerf}
\end{figure}

\textbf{Ref-NeRF} (December 2021) \cite{2022refnerf} was built on mip-NeRF and was designed to better model reflective surfaces. Ref-NeRF parameterized NeRF radiance based on the reflection of the viewing direction about the local normal vector. They modified the density MLP into a directionless MLP, which not only outputs density and the input feature vector of the directional MLP but also diffuse color, specular color, roughness, and surface normal. 
Ref-NeRF performed particularly well on reflective surfaces and is capable of accurately modeling specular reflections and highlights (Fig. \ref{fig:refnerf}).

\textbf{Ray Prior NeRF (RapNeRF)} (May 2022) \cite{2022RaPNeRF} proposed a NeRF model tailored for view extrapolation, contrasting with standard NeRFs that excel at interpolation. RapNeRF introduced Random Ray Casting (RRC), where for a training ray hitting a surface point $\mathbf{v} = \mathbf{o} + t_z \mathbf{d}$, a backward ray is cast from $\mathbf{v}$ toward a new origin $\mathbf{o}'$ with uniformly sampled angular perturbations. Additionally, RapNeRF employed a Ray Atlas (RA) by extracting a rough 3D mesh from a pretrained NeRF and mapping training ray directions onto its vertices. Training began with a baseline NeRF to recover the rough mesh. Subsequently, RRC and RA augmented training rays with a predetermined probability. Evaluations on the Synthetic NeRF dataset \cite{nerf2020_mildenhall} and the MobileObject dataset demonstrated that these augmentations improve view synthesis quality and are adaptable to other NeRF frameworks.

\subsubsection{Depth Supervision and Point Cloud} \label{subsection:ds point cloud}

By using the supervision of expected depth Eq. (\ref{eq:expected depth}) with point clouds acquired from LiDAR or SfM, these models converge faster, converge to higher final quality, and require fewer training views than the baseline NeRF model. Many of these models were also built for or as few-shot/sparse-view NeRF. 

Deng et al. \cite{2022dsnerf} (July 2021) used depth supervision from point clouds with a method named \textbf{Depth-Supervised NeRF (DS-NeRF)}. In addition to color supervision via volume rendering and photometric loss, DS-NeRF also performs depth supervision using sparse point clouds extracted from the training images using COLMAP \cite{2016Colmap}. Depth is modeled as a normal distribution around the depth recorded by the sparse point cloud. A KL divergence term is added to minimize the divergence between the ray’s distribution and this noisy depth distribution (see \cite{2022dsnerf} for details). 

Concurrent to DS-NeRF is a work by \textbf{Roessle et al.} \cite{2022dense_depth_nerf} (April 2021). In this work, the authors used COLMAP to extract a sparse point cloud, which was processed by a Depth Completion Network \cite{2019depth_completionnetwork} to produce depth and uncertainty maps. In addition to the standard volumetric loss, the authors introduced a depth loss based on predicted depth and uncertainty. The model was trained on RGB-D data from ScanNet \cite{2017scannet} and Matterport3D \cite{2017matterport3d} with Gaussian noise added to the depth. 

\textbf{NerfingMVS} (September 2021) \cite{2021nerfingmvs} used multi-view images focusing on depth reconstruction. In NerfingMVS, COLMAP was used to extract sparse depth priors as a point cloud. This was then fed into a pretrained monocular depth network \cite{2019depthnetwork} fine-tuned on the scene to extract a depth map prior. The depth map prior supervised volume sampling by allowing sampling points only at appropriate depths. During volume rendering, the ray was divided into $N$ equal bins, with ray bounds clamped using the depth priors. 

\textbf{PointNeRF} (January 2022) \cite{2022pointnerf} used feature point clouds as an intermediate step before volume rendering. A pretrained 3D CNN \cite{2018mvsnet} generated depth and surface probability $\gamma$ from a cost volume created from training views, producing a dense point cloud. A pretrained 2D CNN \cite{2014vgg} extracted image features from training views. These features populated the point cloud, assigning each point $p_i$ a surface probability $\gamma_i$. Given the input position and view direction, a PointNet \cite{2017pointnet}-like network regressed local density and color, which were then used for volume rendering. Using point cloud features also allowed the model to skip empty spaces, resulting in a speed-up by a factor of 3 over baseline NeRF. 

\subsubsection{Other Geometry Improvement} \label{subsection:geometry improvement}

\textbf{SNeS} (June 2022) \cite{2022snes} improved geometry by learning probable symmetries for partly symmetric and partly hidden in-scene objects through soft symmetry constraints on geometry and material properties.

\textbf{S$^3$-NeRF} (October 2022) \cite{2022s3nerf} used shadow and shading cues to infer scene geometry and enabled single-image NeRF training, with a focus on geometry recovery. S$^3$-NeRF employed a UNISURF-based occupancy field 3D representation instead of density, a modified physics-based rendering equation, and occupancy-based shadow calculation as key implementation differences. The method achieved excellent depth map and surface normal reconstruction from single images on both synthetic and real-world datasets.


\subsection{Improvements to Training and Inference Speed} \label{sec:speed}

In the original implementation by Mildenhall et al. \cite{nerf2020_mildenhall}, hierarchical rendering was employed to improve computational efficiency. Two networks represented the scene: a coarse network and a fine network. The coarse network’s output guided sampling point selection for the fine network, preventing dense sampling at fine scales. In subsequent works over the next two years, most efforts to accelerate NeRF training and inference fall broadly into two categories.

\begin{enumerate}

\item Models in the first category train, precompute, and store NeRF MLP evaluations into more accessible data structures. This improves inference speed significantly but does not affect training time. We refer to these as baked models.

\item The second category comprises non-baked models, which include various innovations. A common approach is to learn separate scene features from MLP parameters via a hybrid representation. This enables smaller MLPs, improving both training and inference speed at the cost of increased memory. Pushing this further, some methods (see Sec. \ref{sec:no mlp}) omit neural networks entirely and use purely explicit scene representations. While not strictly NeRF models, we include them here due to their relevance and similarity to NeRF.

\end{enumerate}
Other techniques include ray termination, which stops sampling when accumulated transmittance nears zero, space skipping, and hierarchical sampling using coarse and fine MLPs as in the original NeRF paper. These methods are often combined with per-paper innovations to further improve training and inference speed.

Hybrid and explicit scene representation methods are closely related to baked methods, since scene features are directly optimized within accessible data structures. However, the baked versus non-baked distinction was popular between 2020 and 2022; as such, we organize this section following that convention.

\subsubsection{Baked}

A model by Hedman et al. \cite{2021sNeRG} (July 2021) stored a precomputed NeRF on a sparse voxel grid. The method, called \textbf{Sparse Neural Voxel Grid (SNeRG)}, stored precomputed diffuse color, density, and feature vectors on the sparse voxel grid in a process sometimes referred to as ``baking". During evaluation, an MLP produced specular color, which, combined with alpha compositing of the specular color along the ray, yielded the final pixel color. The method was approximately 3000 times faster than the original NeRF implementation, with speed comparable to PlenOctree.

Concurrently, the \textbf{PlenOctree} \cite{nerf2021_plenoctrees} approach by Yu et al. (March 2021) achieved inference times approximately 3000 times faster than the original NeRF implementation. The authors trained a spherical harmonic NeRF (NeRF-SH), which predicted spherical harmonic coefficients of the color function instead of directly predicting color values. They constructed an octree of precomputed spherical harmonic (SH) coefficients derived from the MLP’s colors. During octree construction, the scene was voxelized, and voxels with low transmissivity were eliminated. This procedure can also be applied to standard NeRF models by performing Monte Carlo estimations of the spherical harmonic components. PlenOctrees could be further optimized using initial training images via a fast fine-tuning procedure relative to NeRF training. Notably, the Gaussian Splatting implementation of spherical harmonics color is directly adapted from PlenOctree.

In \textbf{FastNeRF} (March 2021) \cite{2021fastNerf}, Garbin et al. factorized the color function $\mathbf{c}$ as the inner product of outputs from two MLPs: a position-dependent MLP that also predicts density $\sigma$, and a direction-dependent MLP. This decomposition enabled FastNeRF to efficiently cache color and density evaluations on a dense scene grid, resulting in inference speedups exceeding 3000 times. The method also leveraged hardware-accelerated ray tracing \cite{2010optix}, which skipped empty space and terminated rays once transmittance saturation was reached.

Reiser et al. (May 2021) \cite{nerf2021_kilonerf} improved the baseline NeRF by introducing \textbf{KiloNeRF}, which divided the scene into thousands of cells and trained independent MLPs to predict color and density for each cell. These small MLPs were trained using knowledge distillation from a large pretrained teacher MLP, a process closely related to baking. The method also employed early ray termination and empty space skipping. These two techniques alone sped up the baseline NeRF’s rendering by a factor of 71. Further splitting the baseline MLP into thousands of smaller MLPs improved rendering speed by a factor of 36, resulting in an overall 2000-fold acceleration.

A paper by Sun et al. \cite{2022voxeldirect} (November 2021) also explored this topic. The authors directly optimized a voxel grid of scalars for density. However, instead of using spherical harmonic coefficients, they used 12- and 24-dimensional features and a small, shallow decoding MLP in a hybrid representation approach. The authors used a sampling strategy analogous to the coarse-fine sampling of the original NeRF paper by training a coarse voxel grid first and then a fine voxel grid based on the geometry of the coarse grid. The model was named \textbf{Direct Voxel Grid Optimization (DVGO)}, which outperformed the baseline NeRF (1–2 days of training) with only 15 minutes of training on the Synthetic NeRF dataset. 

The \textbf{Fourier PlenOctree} \cite{2022fourierplenoct} approach was proposed by Wang et al. in February 2022. It was designed for human silhouette rendering, utilizing the domain-specific technique of Shape-From-Silhouette. The method also draws inspiration from generalizable image-conditioned NeRFs such as \cite{2021mvsnerf} and \cite{nerf2021pixelnerf}. Initially, a coarse visual hull was constructed using sparse views predicted from a generalized NeRF and Shape-From-Silhouette. Colors and densities were then densely sampled inside this hull and stored on a coarse PlenOctree. Dense views were sampled from the PlenOctree, with transmissivity thresholding applied to eliminate most empty points. For the remaining points, new leaf densities and spherical harmonic (SH) color coefficients were generated, and the PlenOctree was updated. A Fourier Transform MLP was subsequently used to extract Fourier coefficients of the density and SH color coefficients, which were fed into an inverse discrete Fourier transform to restore the SH coefficients and density. 

The \textbf{MobileNeRF} (June 2022) \cite{2022mobilenerf} framework trained a NeRF-like model based on a polygonal mesh with color, feature, and opacity MLPs attached to each mesh vertex. Alpha values were discretized, and features were super-sampled for anti-aliasing. During rendering, the mesh with associated features and opacities was rasterized according to the viewing position, and a small MLP was used to shade each pixel. The method demonstrated a speed approximately 10 times faster than SNeRG \cite{2021sNeRG}.

The \textbf{EfficientNeRF} (July 2022) \cite{2022efficientnerf} was based on PlenOctree \cite{nerf2021_plenoctrees}, choosing to use spherical harmonics and cache the trained scene in a tree. However, it introduced several improvements. Most importantly, EfficientNeRF improved training speed by using a momentum density voxel grid to store predicted density via exponential weighted average updates. During the coarse sampling stage, the grid was used to discard sampling points with zero density. During the fine sampling stage, a pivot system was also employed to speed up volume rendering. Pivot points were defined as points $\V{x}_i$ for which $T_i \alpha_i > \epsilon$ where $\epsilon$ was a predefined threshold, and $T_i$ and $\alpha_i$ were the transmittance and alpha values as defined in Eq. (\ref{eq:ray_discrete}) and Eq. (\ref{eq:alpha}). During fine sampling, only points near the pivot points were considered. These two improvements sped up training time by a factor of 8 over the baseline NeRF \cite{nerf2020_mildenhall}. The authors then cached the trained scene into a NeRF tree, resulting in rendering speeds comparable to FastNeRF \cite{2021fastNerf} and exceeding that of baseline NeRF by thousands-fold.

\textbf{R2L} (March 2022) \cite{2022r2l} distilled neural radiance fields into neural light fields via a deep residual MLP. This architecture improved rendering efficiency without requiring data beyond 2D images. Trained by distilling from a pre-trained NeRF, R2L achieved a 26–35× reduction in FLOPs and a 28–31× speedup in wall-clock time while surpassing NeRF and other efficient synthesis methods in visual quality across synthetic and real-world scenes.


\subsubsection{Non-Baked}
A popular early re-implementation of the original NeRF in JAX \cite{jax2018github}, called \textbf{JaxNeRF} \cite{jaxnerf2020github} (December 2020), was often used as a benchmark comparison for early works seeking to improve on training and rendering speed. This model was slightly faster and more suited for distributed computing than the original TensorFlow implementation of NeRF.

In \textbf{Neural Sparse Voxel Fields (NSVF)} (July 2020), Liu et al. \cite{nerf2020_NSVF} developed a voxel-based NeRF model that models the scene as a set of radiance fields bounded by voxels. Feature representations were obtained by interpolating learnable features stored at voxel vertices, which were then processed by a shared MLP that computed $\sigma$ and $\mathbf{c}$. NSVF used a sparse voxel intersection-based point sampling for rays, which was much more efficient than dense sampling or the hierarchical two-step approach of Mildenhall et al. \cite{nerf2020_mildenhall}. However, this approach was more memory-intensive due to storing feature vectors on a potentially dense voxel grid.

\textbf{AutoInt} (December 2020) \cite{2021autoint} approximated the volume rendering step. By separating the discrete volume rendering equation Eq. (\ref{eq:ray_discrete}) piecewise, they developed AutoInt, which trained the MLP $\Phi_\theta$ via its gradient networks $\Psi_\theta^i$. These gradient networks shared internal parameters with and were used to reassemble the integral network $\Phi_\theta$. This approach allowed the rendering step to use far fewer samples, resulting in a tenfold speed-up over the baseline NeRF with only a slight decrease in quality.

\textbf{Recursive‑NeRF} (May 2021) \cite{2022recursive} adapts its representational capacity according to local scene complexity. The method begins with a lightweight neural network that predicts color, volumetric density, and an uncertainty score for each queried spatial point. Points with low uncertainty terminate early. Points with high uncertainty are forwarded to deeper sub-networks branching out dynamically, allowing complex areas to be modelled by richer networks while simple regions require less computation. This early-exit and branching scheme significantly reduces computation compared to standard NeRF: Representing Scenes as Neural Radiance Fields, while maintaining high visual fidelity across scenes.

\textbf{Light Field Networks (LFNs)} (June 2021) \cite{2021lfn} presented a novel neural representation that mapped camera rays directly to radiance in 4D light space, bypassing traditional volumetric queries. This enabled real-time rendering with significantly reduced memory usage and rendering speed improvements by several orders of magnitude. By parameterizing rays using 6D Plücker coordinates, LFNs supported continuous 360° scene representation and encoded both appearance and geometry, from which sparse depth maps could be analytically derived. Although lacking inherent multi-view consistency, LFNs addressed this via a meta-learning framework that enabled light field reconstruction from sparse 2D inputs.

\textbf{Deterministic Integration for Volume Rendering (DIVeR)} (November 2021) \cite{2022DIVeR} took inspiration from NSVF \cite{nerf2020_NSVF} by jointly optimizing a feature voxel grid and a decoder MLP while applying sparsity regularization and voxel culling. However, they innovated the volume rendering process by performing deterministic ray sampling on the voxel grid, which produced an integrated feature for each ray interval—defined by the intersection of the ray with a specific voxel. This feature was then decoded by an MLP to produce the density and color of the ray interval, effectively reversing the usual order of volume sampling and MLP evaluation found in NeRF methods. 
The DIVeR outperformed methods such as PlenOctrees \cite{nerf2021_plenoctrees}, FastNeRF \cite{2021fastNerf}, and KiloNeRF \cite{nerf2021_kilonerf} in terms of quality, at a comparable rendering speed.


\textbf{Instant-Neural Graphics Primitives (INGP)} (January 2022) \cite{nerf2022_ngp} greatly improved NeRF model training and inference speed. The authors proposed a learned parametric multi-resolution hash encoding that was trained simultaneously with the NeRF model MLPs. They also employed advanced ray marching techniques, including exponential stepping, empty space skipping, and sample compaction. This new positional encoding combined with a highly optimized MLP implementation significantly accelerated training and inference, while also enhancing the scene reconstruction accuracy of the resulting NeRF model. Within seconds of training, the method achieved results comparable to hours of training in previous NeRF models.


\subsection{Few Shot/Sparse Training View NeRF} \label{sec:sparse}
The baseline NeRF requires dense multi-view images with known camera poses for each scene. A common failure case for the baseline NeRF is if the training views are not varied enough or the samples are not varied enough in pose. This leads to overfitting to individual views and nonsensical scene geometry. However, a family of NeRF methods leverages pretrained image feature extraction networks, usually a pretrained Convolutional Neural Network (CNN) such as \cite{2012alexnet}\cite{resnet}, to greatly reduce the number of training samples required for successful NeRF training. Some authors call this process ``deep image feature conditioning". Certain methods \cite{2022pointnerf} also used depth/3D geometry supervision to this effect (Sec. \ref{subsection:ds point cloud}). These models often have lower training time compared to baseline NeRF models. Certain previously mentioned methods that regularize or improve geometry also reduce the training view requirement. Methods such as \cite{2022dsnerf}, \cite{2022dense_depth_nerf}, and \cite{2022s3nerf} from section \ref{subsection:ds point cloud} and section \ref{subsection:geometry improvement} approach the sparse view problem by using point clouds for depth supervision. 

In \textbf{pixelNeRF} (December 2020) \cite{nerf2021pixelnerf}, Yu et al. used the pretrained layers of a Convolutional Neural Network (and bilinear interpolation) to extract image features. Camera rays used in NeRF were then projected onto 
the image plane, and the image features were extracted for each query point. The features, view direction, and query points were then passed onto the NeRF MLP, which produced density and color. \textbf{General Radiance Field (GRF)} \cite{nerf2021GRF} (Oct 2020) by Trevithick et al. took a similar approach, with the key difference being that GRF operated in canonical space as opposed to view-space for pixelNeRF. 

\textbf{MVSNeRF} (March 2021) \cite{2021mvsnerf} took a slightly different approach. It extracted 2D image features using a pretrained CNN. These 2D features were mapped to a 3D voxelized cost volume through plane sweeping and a variance-based cost. A pretrained 3D CNN then produced a 3D neural encoding volume, which generated per-point latent codes by interpolation. During volume rendering, the NeRF MLP used these latent features along with point coordinates and viewing direction to predict density and color. The training jointly optimized the 3D feature volume and the NeRF MLP. On the DTU dataset, MVSNeRF reached results comparable to hours of baseline NeRF training within 15 minutes.

\textbf{DietNeRF} (June 2021) \cite{2021dietNeRF} introduced the semantic consistency loss $L_{sc}$ based on image features extracted from CLIP-ViT \cite{2021clipvit}, in addition to the standard photometric loss.
\begin{equation}
    L_{sc} = \frac{\lambda}{2} || \phi(I) - \phi(\hat{I})||^2_2
\end{equation}
where $\phi$ performs the CLIP-ViT feature extraction on training image $I$ and rendered image $\hat{I}$. This is reduced to a cosine similarity loss for normalized feature vectors (eq. 5 in \cite{2021dietNeRF}). DietNeRF was benchmarked on a subsampled NeRF synthetic dataset \cite{nerf2020_mildenhall}, and DTU dataset \cite{2014DTUdataset}. The best-performing method for single-view novel synthesis was a pixelNeRF \cite{nerf2021pixelnerf} model fine-tuned using the semantic consistency loss of DietNeRF.

The \textbf{Neural Rays (NeuRay)} approach, by Liu et al. \cite{2022neuralray} (July 2021) also used a cost volume. From all input views, the authors estimated cost volumes (or depth maps) using multi-view stereo algorithms. From these, a CNN is used to create feature maps $G$. During volume rendering, from these features, both visibility and local features are extracted and processed using MLPs to extract color and alpha. The visibility is computed as a cumulative density function written as a weighted sum of sigmoid functions.

NeuRay generalizes well to new scenes and can be further fine-tuned to exceed the performance of the baseline NeRF model. 

\textbf{GeoNeRF} (November 2021) \cite{2022geonerf} extracted 2D image features from every view using a pretrained Feature Pyramid Network. This method then constructed a cascading 3D cost volume using plane sweeping. From these two feature representations, for each of the $N$ query points along a ray, one view-independent and multiple view-dependent feature tokens were extracted. These were refined using a Transformer \cite{vaswani2017attention}. Then, the $N$ view-independent tokens were refined through an AutoEncoder, which returned the $N$ densities along the ray. The $N$ sets of view-dependent tokens were each fed into an MLP that extracted color. These networks could all be pretrained and generalized well to new scenes, as shown by the authors. Moreover, they could be fine-tuned per scene, achieving great results on the DTU \cite{2014DTUdataset}, NeRF synthetic \cite{nerf2020_mildenhall}, and LLF Forward Facing \cite{2019llf_forwardfacingdataset} datasets, outperforming methods such as pixelNeRF \cite{nerf2021pixelnerf} and MVSNeRF \cite{2021mvsnerf}.

Concurrent to GeoNeRF is \textbf{LOLNeRF} (November 2021) \cite{2022lolnerf}, which was capable of single-shot view synthesis of human faces. It was built similarly to $\pi$-GAN \cite{2021pigan} but used Generative Latent Optimization \cite{2017GLO} instead of adversarial training \cite{2014gan}.

\begin{figure*}[h!] 
\centering
\includegraphics[width=0.9\textwidth]{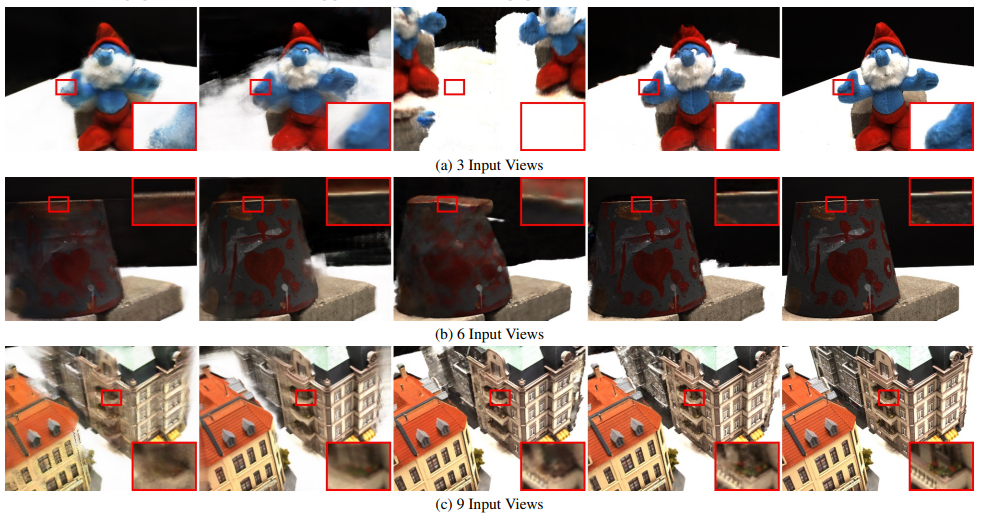}
\caption{Visualization of RegNeRF results on the DTU dataset under sparse view conditions with a) 3, b) 6, and c) 9 input views. Left to right: PixelNeRF \cite{nerf2021pixelnerf}, MVSNeRF \cite{2021mvsnerf}, RegNeRF \cite{2022regnerf}, and ground truth. ©2022 IEEE} \label{fig:regnerf}
\end{figure*}

\textbf{RegNeRF} (December 2021) \cite{2022regnerf} aimed to solve the problem of NeRF training with sparse input views. Unlike most other methods that approached this task by using image features from pretrained networks as a prior for conditioning NeRF volume rendering, RegNeRF employed additional depth and color regularization.
The depth smoothness encourages world geometry to be piecewise smooth, and is defined as 
\begin{equation}
    L_{ds} = \sum_{\V{r}} \sum_{ij}^{S_{patch-1}} (d(\V{r_{ij}})-d(\V{r_{i+1j}}))^2+(d(\V{r_{ij}})-d(\V{r_{ij+1}}))^2
\end{equation}
where $d(\V{r_{ij}})$ refers to the expected depth of a ray through pixel $ij$ of a patch of size $S_{patch}$, from a randomly sampled unobserved viewpoint. Color regularization was also used by estimating and maximizing the likelihood of rendered patches. This was done by training a Normalizing Flow model such as RealNVP \cite{2016realnvp} on a varied unposed dataset such as JFT-300M \cite{2017jftdataset}, then estimating and maximizing the log-likelihood of rendered patches. Let $\phi$ be a learned bijection from patches to $R^d$ where $d=S_{patch}\cdot S_{patch} \cdot3$. The color regularization loss is defined as 
\begin{equation}
    L_{NLL} = \sum_{\V{r}} - \log p_z (\phi(P_{r}))
\end{equation}
where $P_r$ is the predicted RGB color patch with center pixel at $\V{r}$, and $-\log p_z$ is the negative log-likelihood with Gaussian $p_z$. In addition, RegNeRF used sampled space annealing, which attempted to fix divergent NeRF modes with high density at the ray origin at the start of training. This was done by limiting the range of sample points to within a small volume defined for all input images before extending to the full scene. Mip-NeRF \cite{2021mipnerf} was used as the backbone NeRF model for these regularization and sampling techniques. 
The model was tested on DTU \cite{2014DTUdataset} and LLFF \cite{2019llf_forwardfacingdataset} datasets and outperformed models such as PixelNeRF \cite{nerf2021pixelnerf}, SRF \cite{2021SRF}, and MVSNeRF \cite{2021mvsnerf}. RegNeRF, which did not require pretraining, achieved comparable performance to these models, which were pre-trained on DTU and fine-tuned per scene, outperforming Mip-NeRF and DietNeRF \cite{2021dietNeRF} under sparse view conditions (see Figure \ref{fig:regnerf}).

\textbf{NeRFusion} (March 2022) \cite{2022nerfusion} also extracted a 3D cost volume from 2D image features extracted from CNN. The volume was then processed by a sparse 3D CNN into a local feature volume. This method performs this step for each frame and then uses a GRU \cite{2014gru} to fuse these local feature volumes into a global feature volume, which was used to condition density and color MLPs. NeRFusion outperformed the baseline NeRF \cite{nerf2020_mildenhall}, NeRFingMVS \cite{2021nerfingmvs}, MVSNeRF \cite{2021mvsnerf} on ScanNet \cite{2017scannet}, DTU \cite{2014DTUdataset}, and NeRF Synthetic \cite{nerf2020_mildenhall} datasets. 

\textbf{AutoRF} (April 2022) \cite{2022autorf} focused on novel view synthesis of objects without background. Given 2D multi-view images, a 3D object detection algorithm was used with panoptic segmentation to extract 3D bounding boxes and object masks. The bounding boxes were used to define Normalized Object Coordinate Spaces, which were used for per-object volume rendering. An encoder CNN was used to extract appearance and shape codes, which were used in the same way as in GRAF \cite{2020graf}. In addition to the standard photometric loss Eq. (\ref{eq:photo-loss}), an additional occupancy loss was defined as 
\begin{equation}
    L_{occ} = \frac{1}{|W_{occ}|} \sum_{u \in W_{occ}} \log (Y_u(1/2 -\alpha)+1/2)
\end{equation}
where $Y$ is the object mask, and $W_{occ}$ is either the set of foreground or background pixels. During test time, the shape codes, appearance codes, and bounding boxes were further refined using the same loss function. 

\textbf{SinNeRF} (April 2022) \cite{2022sinnerf}attempted NeRF scene reconstruction from single images by integrating multiple techniques. They used image warping and known camera intrinsics and poses to create reference depth for depth supervision of unseen views. They employed adversarial training with a CNN discriminator to provide patch-wise texture supervision. Additionally, they used a pretrained ViT to extract global image features from the reference patch and unseen patch, comparing them with an L2 loss term and a global structure prior. SinNeRF outperformed DS-NeRF \cite{2022dsnerf}, PixelNeRF \cite{nerf2021pixelnerf}, and DietNeRF \cite{2021dietNeRF} on the NeRF Synthetic dataset \cite{nerf2020_mildenhall}, the DTU dataset \cite{2014DTUdataset}, and the LLFF Forward Facing dataset \cite{2019llf_forwardfacingdataset}.

As an alternate approach, \textbf{GeoAug} (Oct 2022) \cite{2022geoaug} used data augmentation by rendering (with warping) new training images with new noisy camera poses using DSNeRF \cite{2022dense_depth_nerf} as a baseline and using depth as regularization.
\subsection{Generative and Conditional Models}\label{sec:conditional}

Inspired by advances in generative 2D computer vision, generative NeRF models produce 3D geometry conditioned on text, images, or latent codes. This conditioning enables some degree of scene editing. These models fall broadly into two categories: Generative Adversarial Network-based methods and Diffusion-based methods. Typically, they leverage 2D generative models to create images of the ‘scene,’ which are then used to train the NeRF model. A major challenge before Gaussian Splatting was generating 2D images conditioned on camera pose in a way that maintained 3D consistency. Another persistent issue is the multi-face Janus problem, where generative NeRFs create avatars with multiple faces around the head. This Janus problem remains an active research area even after Gaussian Splatting’s introduction.

Compared to later 2D image generation models based on diffusion and flow-matching, the early GAN-based image generation from the early NeRF era was conditioned on latent code and could not be easily controlled with text and image-based conditioning. Without text and image-based conditioning, latent codes were also used to control aspects of scene editing. 

In \textbf{NeRF-VAE} (January 2021) \cite{2021VaeNerf}, Kosiorek et al. proposed a generative NeRF model that generalized well to out-of-distribution scenes and removed the need to train on each scene from scratch. The NeRF renderer in NeRF-VAE was conditioned on latent code, which was trained using Iterative Amortized Inference \cite{2018iai}\cite{2018iai2} and a ResNet \cite{resnet} encoder. The authors also introduced an attention-based scene function (as opposed to the typical MLP). NeRF-VAE consistently outperformed the baseline NeRF with a low number (5-20) of scene views, but due to lower scene expressivity, it was outperformed by baseline NeRF when a large number of views were available (100+). 

\subsubsection{Generative Adversarial Network-based methods} \label{sec:gan}
Adversarial training is often used for generative and/or latent-conditioned NeRF models. First developed in 2014, Generative Adversarial Networks (GANs) \cite{2014gan} are generative models that employ a generator $G$ which synthesizes images from ``latent code/noise", and a discriminator $D$ which classifies images as ``synthesized" or ``real". The generator seeks to ``trick" the discriminator and make its images indistinguishable from ``real" training images. The discriminator seeks to maximize its classification accuracy. These two networks are trained adversarially, which is the optimization of the following minimax loss/value function,
\begin{equation}
\min_G \max_D \mathbb{E}{x \sim data}[\log D(x)] + \mathbb{E}{z\sim p(z)}[\log(1-D(G(z)))]
\end{equation}
where the generator generates images based on latent code $z$ sampled from some distribution $p(z)$, which the discriminator compares to training image $x$. In GAN-based generative NeRF models, the generator $G$ encompasses all novel-view synthesis steps and is thought of as the NeRF model. The generator in this case also requires an input pose in addition to a latent code. The discriminator $D$ is usually an image classification CNN.
\textbf{GRAF} (July 2020) \cite{2020graf} was the first NeRF model trained adversarially. It paved the way for many later works. A NeRF-based generator was conditioned on latent appearance code $\mathbf{z_a}$ and shape code $\mathbf{z_s}$, and is given by
\begin{equation}
G (\gamma(\mathbf{x}), \gamma(\mathbf{d}), \mathbf{z_s}, \mathbf{z_a}) \rightarrow (\sigma, \mathbf{c}).
\end{equation}
In practice, the shape code, conditioning scene density, was concatenated with the embedded position, which was input to the direction-independent MLP. The appearance code, conditioning scene radiance, was concatenated with the embedded viewing direction, which was input to the direction-dependent MLP. As per baseline NeRF, images were generated via volume sampling. These were then compared using a discriminator CNN for adversarial training. 

Within three months of GRAF, Chan et al. developed \textbf{$\mathbf{\pi}$-GAN} (December 2020) \cite{2021pigan}, which also used a GAN approach to train a conditional NeRF model. The generator was a SIREN-based \cite{2020siren} NeRF volumetric renderer, with sinusoidal activation replacing the standard ReLU activation in the density and color MLPs. $\pi$-GAN outperformed GRAF \cite{2020graf} on standard GAN datasets such as Celeb-A \cite{2016celeba}, CARLA \cite{2017carla}, and Cats \cite{2008cats}.

\textbf{EG3D} (December 2021) \cite{EG3D} uses a novel hybrid tri-plane representation, with features stored on three axis-aligned planes and a small decoder MLP for neural rendering in a GAN framework. The GAN framework is composed of a pose-conditioned StyleGAN2 feature map generator for the tri-plane, a NeRF rendering module converting tri-plane features into low-resolution images, and a super-resolution module. The super-resolved image is then fed into a StyleGAN2 discriminator. The model achieved state-of-the-art results on the FFHQ dataset, producing realistic images and 3D geometry of human faces. 

\textbf{StyleNeRF} (January 2022) \cite{2022stylenerf} is a highly influential work that focuses on 2D image synthesis by using NeRF to bring 3D awareness to the StyleGAN \cite{2019styleGANFFHQ} \cite{2020stylegan2} image synthesis framework. StyleNeRF uses a style-code-conditioned NeRF with an upsampling module as the generator, and a StyleGAN2 \cite{2020stylegan2} discriminator, and introduces a new path regularization term to the StyleGAN optimization objective.

\textbf{Pix2NeRF} (February 2022) \cite{2022pix2nerf} was proposed as an adversarially trained (based on $\pi$-GAN \cite{2021pigan}) NeRF model which could generate NeRF-rendered images given randomly sampled latent codes and poses. In addition to the $\pi$-GAN loss, from which the adversarial architecture is based, the Pix2NeRF loss function also includes the following: 1) a reconstruction loss comparing $z_{predicted}$ and $z_{sampled}$ to ensure consistency of latent space, 2) a reconstruction loss ensuring image reconstruction quality, between $I_{real}$ and $I_{reconstructed}$, where $I_{reconstructed}$ is created by the generator from a $z_{pred}, d_{pred}$ pair produced by the encoder, 3) a conditional adversarial objective which prevents mode collapse towards trivial poses (see original paper for the exact expressions). 

\subsubsection{Jointly Optimized Latent Models}
These models use latent codes as a key aspect of view-synthesis but jointly optimize the latent code with the scene model. The models listed in this section are not generative but instead use latent codes to account for various changeable aspects of the scene. 
In Generative Latent Optimization (GLO) \cite{2017GLO}, a set of randomly sampled latent codes $\{\V{z_1},...,\V{z_n}\}$, usually normally distributed, is paired to a set of images $\{I_1,...,I_n \}$. These latent codes are input to a generator $G$ whose parameters are jointly optimized with the latent code using some reconstruction loss $L$ such as $L_2$. I.e., the optimization is formulated as 
\begin{equation}
    \min_{G,\V{z_i},...,\V{z_n}} \sum_{i=i}^{n} L(G (\V{z_i}, \V{u_i}),I_i)
\end{equation}
where $\V{u_i}$ represents the other inputs not optimized over (needed in NeRF but not necessarily for other models). According to the GLO authors, this method can be thought of as a Discriminator-less GAN. 

One should note that in the 2020–2023 era, many NeRF models use latent codes to condition certain aspects of the scene, such as the appearance and transient embeddings in NeRF-W. These models are typically optimized using GLO. We do not list them in this section unless the latent code is used explicitly for scene editing as a key idea in the paper.

\textbf{Edit-NeRF} (June 2021) \cite{2021editnerf} allowed for scene editing using image conditioning from user input. Edit-NeRF's shape representation was composed of a category-specific shared shape network $F_{share}$ and an instance-specific shape network $F_{inst}$. $F_{inst}$ was conditioned on $\mathbf{z_s}$, whereas $F_{shared}$ was not. In theory, $F_{shared}$ behaved as a deformation field, not unlike \cite{nerf2021_nerfies}. The NeRF editing was formulated as a joint optimization problem of both the NeRF network parameters and the latent codes $\mathbf{z_s, z_a}$, using GLO. 
They then conducted NeRF photometric loss optimization on latent codes, then on the MLP weights, and finally optimized both latent codes and weights jointly. 

Innovating on Edit-NeRF, \textbf{CLIP-NeRF}'s (December 2021) \cite{2022clipnerf} neural radiance field was based on the standard latent-conditioned NeRF, i.e., NeRF models conditioned on shape and appearance latent codes. However, by using Contrastive Language–Image Pre-training (CLIP), CLIP-NeRF could extract from user input text or images the induced latent space displacements by using shape and appearance mapper networks. These displacements could then be used to modify the scene's NeRF representation based on these input text or images. This step allowed for skipping the per-edit latent code optimization used in Edit-NeRF, resulting in a speed-up of a factor of $\sim$8–60, depending on the task. They also used a deformation network, similar to deformable NeRFs \cite{nerf2021_nerfies} (called the \textit{instance-specific network} in Edit-NeRF \cite{2021editnerf}), to help with modifying the scene based on latent space displacement. 
\begin{figure}[h!] 
\centering
\includegraphics[width=0.45\textwidth]{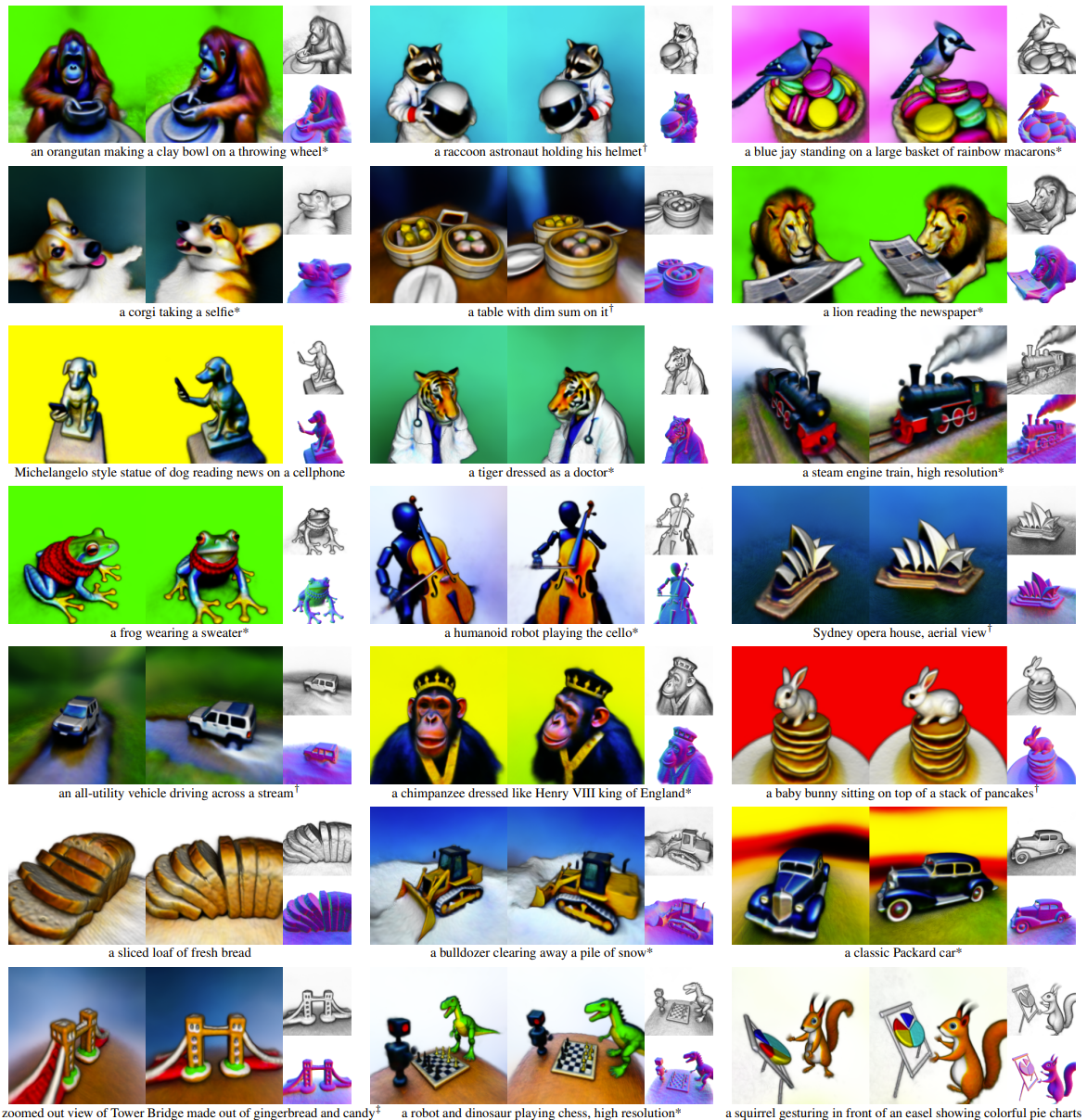}
\caption{Examples of text-to-3D results from Dreamfusion \cite{2022dreamfusion} A pretrained text-to-image diffusion model is used to generate a training set in a NeRF-based pipeline. © Authors \cite{2022dreamfusion}} \label{fig:dreamfusion}
\end{figure}

\subsubsection{Diffusion NeRF Models}

Diffusion models are a family of image generation and editing methods that have attracted widespread attention in 2022 and have largely overtaken GAN methods for 2D image generation post-Gaussian Splatting. Diffusion models are trained using a forward and a reverse diffusion process. The forward diffusion process adds noise to some input image/feature map in some T steps \cite{2015diffusionprobabilistic}. The reverse process is generative and can be used to create images from Gaussian noise \cite{2022imagen}\cite{2022LDM}. Diffusion models offer a high degree of control over image generation by allowing for text and image-based prompting/conditioning using domain-specific encoders.

\textbf{DreamFusion} (September 2022) \cite{2022dreamfusion} was proposed as a text-to-3D diffusion NeRF model. The NeRF model in DreamFusion was trained using images from 2D diffusion models. For each object or scene to be generated, text prompts are input into the diffusion model Imagen \cite{2022imagen}, and a mip-NeRF 360 \cite{2022mip360} based NeRF model was trained from scratch. The text prompt allowed for control over the view of the subject at the diffusion image generation stage, with some prompting using keywords such as ``overhead view", ``front view", and ``back view". A key modification to the NeRF training was that surface color was parameterized by an MLP instead of radiance. Despite impressive results, Imagen images were generated at a 64x64 resolution. As such, the resulting NeRF model lacked the capability to produce finer details. Some results are shown in Figure \ref{fig:dreamfusion}.

In \textbf{Latent-NeRF} (November 2022) \cite{2022ldnerf}, the NeRF model was trained to output 64x64x4 latent features that Stable Diffusion \cite{2022LDM} operated over, which then resulted in 512x512x3 RGB images after a decoder step. The method allowed for text guidance and shape guidance, both for further shape refinement and as a strict shape constraint.

Building upon DreamFusion, \textbf{Magic3D} (November 2022) \cite{2022magic3d} targeted the issues caused by low-resolution diffusion images. The authors used a two-stage coarse-fine approach. In the coarse stage, Magic3D used Instant-NGP \cite{nerf2022_ngp} as the NeRF model trained from images generated from text prompts using the image diffusion model eDiff-I \cite{2022ediffi}. The coarse geometry extracted from Instant-NGP was then placed on a mesh, which was optimized over in the fine stage using images generated with a latent diffusion model \cite{2022LDM}. The authors noted that their method allowed for prompt-based scene editing, personalized text-to-3D generation via conditioning on an image of a subject, and style-guided text-to-3D generation. Their experiments over 397 prompts generated objects, each rated by three users, also showed a preference toward Magic3D over DreamFusion.

\textbf{RealFusion} (February 2023) \cite{2023realfusion} also used some of the same ideas but focused on single-shot scene learning. The underlying diffusion model is Stable Diffusion \cite{2022LDM}, and the underlying NeRF model is Instant-NGP \cite{nerf2022_ngp}. The authors used single-image textual inversion as a substitute for alternate views by augmenting the input 2D image and associating it with a new vocabulary token to optimize the diffusion loss, which ensures the radiance field represents the object in the single-view photography. The 3D scene was then trained in a coarse-to-fine manner using the NeRF photometric loss, which reconstructs the scene.

\textbf{SSDNeRF} (April 2023) \cite{2023ssdnerf} learns generalizable 3D priors through a single-stage 3D latent diffusion model. Unlike many two-stage approaches that separately train autoencoders and diffusion models, often leading to noisy latent representations, SSDNeRF jointly optimizes NeRF and diffusion components end-to-end from multi-view images. This strategy enables robust learning even with sparse-view input. Additionally, the model supports flexible test-time sampling, allowing 3D reconstruction from arbitrary view counts. Experiments on single-object datasets show strong performance across generative and reconstruction tasks, advancing toward a general-purpose 3D learning framework.

In addition to these generative diffusion models, diffusion models are also used for single-view NeRF scene learning via image conditioning (\textbf{NeuralLift-360} (Nov 2022) \cite{2022neurallift}, \textbf{NeRFDi} (December 2022) \cite{2022nerdi}, \textbf{NerfDiff} (February 2023) \cite{nerfdiff}, \textbf{PoseDiff} (Jan 2024) \cite{2024posediff}
), as well as for geometry regularization, (\textbf{DiffusioNeRF} (February 2023) \cite{2023diffusionerf}).


\subsection{Unbounded Scene and Scene Composition} \label{sec:composition}
With attempts using NeRF models in outdoor scenes came a need to separate foreground and background, which may contain views of the sky or the horizon. These outdoor scenes also posed additional challenges in image-by-image variation in lighting and appearance. The models introduced in this section approached this problem using various methods, with many models adapting the latent conditioning via image-by-image appearance codes. Certain models in this research area also perform semantic or instance segmentation to find applications in 3D semantic labeling. We also include a subsection of 4D NeRF methods built for scenes with dynamic objects, which often require some level of semantic understanding separating dynamic objects from static objects

In \textbf{NeRF in the Wild (NeRF-W)}  (August 2020) \cite{nerf2020_nerfw}, Martin-Brualla et al. addressed two key issues of early NeRF models. Real-life photographs of the same scene can contain per-image appearance variations due to lighting conditions, as well as transient objects that differ in each image. The density MLP was kept fixed for all images in a scene. However, NeRF-W conditioned their color MLP on a per-image appearance embedding. Moreover, another MLP conditioned on a per-image transient embedding predicted the color and density functions of transient objects. 

Zhang et al. developed the \textbf{NeRF++} (October 2020) \cite{nerf2020nerf++} model, which was adapted to generate novel views for unbounded scenes by separating the scene using a sphere. The inside of the sphere contained all foreground objects and all fictitious camera views, whereas the background was outside the sphere. The outside of the sphere was then reparameterized with radial inversion. Two separate NeRF models were trained, one for the inside of the sphere and one for the outside. The camera ray integral was also evaluated in two parts. 

\textbf{GIRAFFE} (November 2020) \cite{2021giraffe} was built with a similar approach to NeRF-W, using generative latent codes and separating background and foreground MLPs for scene composition. GIRAFFE was based on GRAF, a previous model used for generative scene modeling. The framework assigned to each object in the scene its own neural feature field MLP, which produced a scalar density and a deep feature vector replacing color. These MLPs, with shared architecture and weights, took as input shape and appearance latent vectors as well as an input pose. 
The scene was then composed using a density-weighted sum of features. A small 2D feature map was then created from this 3D volume feature field using volume rendering, which was fed into an upsampling CNN to produce an image. GIRAFFE performed adversarial training using this synthesized image and a 2D CNN discriminator. The resulting model had a disentangled latent space, allowing for fine control over scene generation.

\textbf{Fig-NeRF} (April 2021) \cite{2021fig} also took on scene composition but focused on object interpolation and amodal segmentation. They used two separate NeRF models, one for the foreground and one for the background. Their foreground model was the deformable Nerfies model \cite{nerf2021_nerfies}. Their background model was an appearance NeRF conditioned on latent codes. They used two photometric losses, one for the foreground and one for the background.
Fig-NeRF achieved good results for amodal segmentation and object interpolation, on datasets such as ShapeNet \cite{2015shapenet}, Gelato \cite{2020gelato}, and Objectron \cite{2021objectron}.

\textbf{Yang et al.} (September 2021) \cite{2021YangObjectComposition} created \textbf{Object-NeRF}, a composition model that can separate and edit objects within the scene. They used a hybrid voxel-based approach \cite{nerf2020_NSVF}, creating a voxel grid of features that is jointly optimized with MLP parameters. They used two different NeRFs, one for objects and one for the scene, both of which were conditioned on interpolated voxel features. The object NeRF was further conditioned on a set of object activation latent codes. Their method was trained and evaluated on ScanNet \cite{2017scannet} as well as an in-house ToyDesk dataset with instance segmentation labels. They incorporated segmentation labels with a mask loss term, identifying each in-scene object.

\begin{figure}[htpb] 
\centering
\includegraphics[width=0.45\textwidth]{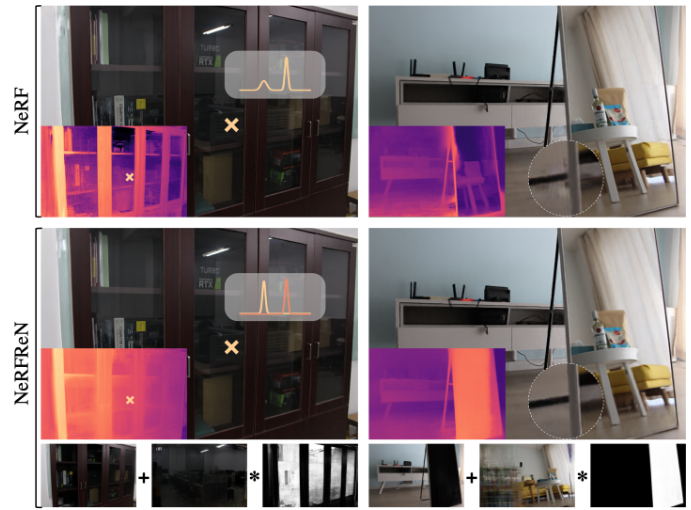}
\caption{NeRFRen \cite{2022nerfren} comparison in a scene with reflective surfaces. NeRFRen is capable of accurately reconstructing the depth (and geometry ) of reflective surfaces such as glasses, a common failure case for neural rendering standard approaches. ©2022 IEEE}  \label{fig:nerfren}
\end{figure}

\textbf{NeRFReN} (November 2021) \cite{2022nerfren} addressed the problem of reflective surfaces in NeRF view synthesis. The authors separated the radiance field into two components, transmitted $(\sigma^t,\V{c}^t)$ and reflected $(\sigma^r,\V{c}^r)$, with the final pixel value given by
\begin{equation}
    I = I_t + \beta I_r
\end{equation}
where $\beta$ is the reflection fraction given by the geometry of the transmitted radiance field as 
\begin{equation}
    \beta = \sum_i T_{\sigma_i^t} (1-\exp(-\sigma_i^t \delta_i))\alpha_i.
\end{equation}
$T_{\sigma_i^t}$ is given by Eq. (\ref{eq:transmisivity}), and $\alpha_i$ by
Eq. (\ref{eq:alpha}). 
In addition to the standard photometric loss, the authors used a depth smoothness $L_d$ (eq. 8 in \cite{2022nerfren}) loss to encourage the transmitted radiance field to produce the correct geometry. Likewise, a bidirectional depth consistency loss $L_{bdc}$ (eq. 10 in \cite{2022nerfren}) was used for the reflected radiance field. NeRFReN was able to render reflective surfaces on the authors' RFFR dataset, outperforming benchmark methods such as baseline NeRF \cite{nerf2020_mildenhall}, and NerfingMVS \cite{2021nerfingmvs}, as well as ablation models. The method was shown to support scene editing via reflection removal and reflection substitution. 

\subsubsection{4D NeRF and Dynamic Scenes}\label{sec:4D}
{\textbf{D-NeRF} (November 2020) \cite{2021dnerf} authors extend the NeRF to handle dynamic scenes by introducing time $t$ as an additional input and splitting the representation into two learned neural modules. One module, the deformation network, maps a point in 4D space-time into a canonical “shape” space via a learned displacement field. The second, the canonical radiance network, then predicts density and view-dependent color from the canonical coordinate $x'$ and viewing direction $d$. During rendering, they sample along rays as usual but apply the deformation transform before querying the canonical field; this lets the model represent non-rigid, articulated, and rigid motions from a monocular moving camera.

\textbf{DeVRF} (May 2022) \cite{2022devrf} proposes adapting the NeRF-like volume rendering frameworks with explicit 3D representation (such as Plenoxels and TensoRF) to dynamic scenes by using an explicit voxel-based canonical 3D volumetric grid together with a 4D deformation field (space + time) superimposed onto that grid. It employs a static-to-dynamic learning paradigm: first learn the static scene representation from multi-view stills, then learn the deformation from a few-view dynamic sequence, allowing much faster convergence than fully implicit networks.

Robust Dynamic Radiance Fields \textbf{(RoDynRF)} (January 2023) \cite{2023rodynrf} enables dynamic scene reconstruction from monocular videos without known camera poses by jointly optimizing intrinsics, extrinsics, and both static and dynamic radiance fields. It introduces a two-branch architecture separating background and moving regions, uses motion masks to exclude dynamic pixels during pose estimation, and employs a coarse-to-fine voxel optimization scheme for stability. A time-conditioned deformation network models non-rigid motion, allowing robust recovery of dynamic geometry and appearance from unposed, in-the-wild video data.

\textbf{HexPlane} (January 2023) \cite{2023hexplane} and \textbf{K-Planes} (January 2023) \cite{2023kplanes} introduce a hybrid 4D representation for dynamic scenes by using six learned feature-planes that span pairs of spacetime axes, fusing per-point features from each plane. Both methods use a decoder; Hexplanes decode via a small MLP with volume-rendering supervision, whereas K-planes also provide the option of using a linear decoder. These decoders are used to produce local color and density for NeRF-like volume rendering. The creation of to-be-decoded features from the 6 feature planes also slightly differs in both methods, resulting in minor differences in performance. The K-Plane implementation is more general, allowing for $ n$-dimensional representation (as opposed to 4D). By using this hybrid scheme, both methods result in a compression with respect to a 4D voxel-based representation, as well as a significant speed up with respect to fully implicit NeRF baselines.

\textbf{MAV3D} (January 2023) \cite{2023mav3d} synthesizes time-varying 3D scenes directly from text by optimizing a 4D neural radiance field under guidance from a pretrained text-to-video diffusion model. It introduces a hexplane-based 4D representation for efficient spacetime encoding, a multi-stage static-to-dynamic training scheme with temporal regularization, and a data-free supervision strategy using only text and video priors to create dynamic 3D content without any 4D datasets.

\textbf{NeRF-DS} (March 2023) \cite{2023nerfds} extends NeRF to capture dynamic, reflective objects by conditioning radiance on surface orientation and spatial location, and by using an object mask to maintain temporal consistency under changing reflections. Methodologically, it employs a dual-network design like D-NeRF \cite{2021dnerf}: a deformation field maps time-varying coordinates into a canonical space, while a specularity-aware radiance field (improving upon D-NeRF) jointly predicts view-dependent appearance and reflection behavior, enabling faithful modeling of non-Lambertian motion.

Methods built for animating human faces (HyperNeRF \cite{nerf2021_hypernerf}), bodies (PREF \cite{2022pref}, A-NeRF \cite{2021anerf}) , or articulated objects (LISA \cite{2022lisa}), are discussed separately in Section \ref{sec:42}.

\subsection{Pose Estimation} \label{sec:pose}
NeRF models require both input images and camera poses to train. In the original 2020 paper, unknown poses were acquired by the COLMAP library \cite{2016Colmap}, which was also often used in many subsequent NeRF models when camera poses were not provided. Typically, building models that perform both pose estimation and implicit scene representation with NeRF is formulated as an offline structure from motion (SfM) problem. In these cases, Bundle Adjustment (BA) is often used to jointly optimize the poses and the model. However, some methods also formulate this as an online simultaneous localization and mapping (SLAM) problem. 

\textbf{iNeRF} (December 2020) \cite{2021inerf} formulated pose reconstruction as an inverse problem. Given a pre-trained NeRF, using the photo-metric loss Eq. \ref{eq:photo-loss}, Yen-Chen et al. optimized the pose instead of the network parameters. The authors used an interest-point detector and performed interest region-based sampling. The authors also performed semi-supervised experiments, where they used iNeRF pose estimation on unposed training images to augment the NeRF training set, and further train the forward NeRF. This semi-supervision was shown by the author to reduce the requirement of posed photos from the forward NeRF by 25 \%.

\textbf{NeRF--} (February 2021) \cite{nerf2021_nerf--} jointly estimated NeRF model parameters and camera parameters. This allowed for the model to construct radiance fields and synthesize novel view images in an end-to-end manner. NeRF-- overall achieved comparable results to using COLMAP with the 2020 NeRF model in terms of view synthesis. However, due to limitations with pose initialization, NeRF-- was most suited for front-facing scenes, and struggled with rotational motion and object tracking movements. 

Concurrent to NeRF-- was the \textbf{Bundle-Adjusted Neural Radiance Field (BARF)} (April 2021) \cite{2021barf}, which also jointly estimated poses alongside the training of the neural radiance field. BARF also used a coarse-to-fine registration by adaptively masking the positional encoding, similar to the technique used in Nerfies \cite{nerf2021_nerfies}. Overall, BARF results exceeded those of NeRF-- on the LLFF forward-facing scenes dataset with unknown camera poses by 1.49 PSNR averaged over the eight scenes, and outperformed COLMAP registered baseline NeRF by 0.45 PSNR. Both BARF and NeRF-- used naive dense ray sampling for simplicity. 

Jeong et al. introduced a self-calibrating joint optimization model for NeRF \textbf{(SCNeRF)} (August 2021) \cite{nerf2021self_calibrating}. Their camera calibration model can not only optimize unknown poses, but also camera intrinsic parameters for non-linear camera models such as fish-eye lens models. By using curriculum learning, they gradually introduce the nonlinear camera/noise parameters to the joint optimization. This camera optimization model was also modular and could be easily used with different NeRF models. The method outperformed BARF \cite{2021barf} on LLFF scenes \cite{2019llf_forwardfacingdataset}.

\textbf{GNeRF} (March 2021) \cite{2021Gnerf}, a different type of approach by Meng et al., used pose as a generative latent code. GNeRF first obtains coarse camera poses and a radiance field with adversarial training. This is done by using a generator that takes a randomly sampled pose and synthesizes a view using NeRF-style rendering. Then, a discriminator compares the rendered view with the training image. An inversion network then takes the generated image and outputs a pose, which is compared to the sampled pose. This results in a coarse image-pose pairing. The images and poses are then jointly refined via a photometric loss in a hybrid optimization scheme. GNeRF was slightly outperformed by COLMAP-based NeRF on the Synthetic-NeRF dataset and outperformed COLMAP-based NeRF on the DTU dataset.

\textbf{GARF} (April 2022) \cite{2022garf} used Gaussian activations as an effective alternative to positional encoding in NeRF, in conjunction with bundle adjustment for pose estimation. The authors showed that GARF can successfully recover scene representations from unknown camera poses, even in challenging scenes with low-textured regions, making it suitable for real-world applications. 

\subsubsection{NeRF and SLAM}

Sucar et al. introduced the first NeRF-based dense online SLAM model named \textbf{iMAP} (March 2021) \cite{2021imap}. The model jointly optimizes the camera pose and the implicit scene representation in the form of a NeRF model. They used an iterative two-step approach of tracking, which is pose optimization with respect to NeRF, and mapping, which is bundle adjustment for joint optimization of pose and NeRF model parameters. iMAP achieved a pose tracking speed close to the camera framerate by running the much faster tracking step in parallel with the mapping process. However, when combined with the much slower NeRF-based mapping step iMAP is far from achieving real-time performance. iMAP also used keyframe selection by optimizing the scene on a sparse and incrementally selected set of images.

Building on iMAP, \textbf{NICE-SLAM} (December 2021) \cite{2022niceslam} improved various aspects such as keyframe selection and NeRF architecture. Specifically, they used a hierarchical grid-based representation of the scene geometry, which was able to fill in gaps in iMAP reconstruction of large-scale unobserved scene features like walls and floors in certain scenes. NICE-SLAM achieved lower pose estimation errors and better scene reconstruction results than iMAP. NICE-SLAM also used approximately one-quarter of the FLOPs of iMAP, one-third of the tracking time, and half of the mapping time but is still far from achieving real-time performance).

\textbf{NeRF-SLAM} (October 2022) \cite{2022nerfSLAM} improved on the existing NeRF-based SLAM approach by using Instant-NGP \cite{nerf2022_ngp} as its mapping module's NeRF model, in conjunction with a state-of-the-art SLAM pipeline, greatly exceeding previous benchmarks on the Replica dataset \cite{2019replica}. 

\textbf{Vox-Fusion} (October 2022) \cite{2022voxfusion} introduces a hybrid SLAM system that combines a voxel-based neural implicit surface representation with an octree structure for dynamic scene expansion. It encodes and optimizes each scene voxel implicitly, supports incremental mapping without prior environment knowledge, leverages a multi-process framework for heightened speed, and demonstrates improved accuracy and completeness compared to prior methods.

\textbf{NICER-SLAM} (February 2023) \cite{2023nicerslam} is an end-to-end dense SLAM system that performs simultaneous tracking and mapping using only RGB inputs, improving on NICE-SLAM \cite{2022niceslam}. It introduces a hierarchical neural implicit representation based on signed distance functions (SDFs), enabling detailed 3D geometry and photorealistic novel view synthesis. The system leverages monocular geometric cues, optical flow, and a warping loss to guide optimization without depth supervision. Additionally, it proposes a locally adaptive SDF-to-density transformation tailored for indoor scene dynamics.

\subsection{Adjacent Methods for Neural Rendering} \label{sec:adjacent}

\subsubsection{Explicit Representation and Fast MLP-less Volume Rendering} \label{sec:no mlp}

\textbf{Plenoxels} (December 2021) \cite{2021plenoxels} followed in Plenoctree's footsteps by voxelizing the scene and storing a scalar for density and spherical harmonics coefficients for direction-dependent color. However, surprisingly, Plenoxel skipped the MLP training entirely and instead fit these features directly on the voxel grid. They also obtained comparable results to NeRF++ and JaxNeRF, with training times faster by a factor of a few hundred. These results showed that the primary contribution of NeRF models is the volumetric rendering of new views given point-wise densities and color, not the density and color MLPs themselves. \textbf{HDR-Plenoxels} \cite{2022hdrplenoxel} (August 2022) adapted this idea to HDR images by learning 3D High Dynamic Range radiance fields, scene geometry, and various camera settings from Low Dynamic Range images.


\textbf{TensoRF} (March 2022) \cite{2022tensorf} stored a scalar density and a vector feature (can work with SH coefficients, or features to be decoded via MLP) as factorized tensors. These were initially represented as a rank 3 tensor $T_\sigma \in R^{H\times W \times D}$ and a rank 4 tensor $T_c \in R^{H\times W \times D \times C}$, where $H,W,D$ are the height, width, and depth resolution of the voxel grid, and $C$ is channel dimension. The authors then used two factorization schemes: CANDECOMP-PARAFAC (CP), which factorized the tensors as pure vector outer products, and Vector Matrix (VM), which factorized the tensors as vector/matrix outer products. These factorizations decreased the memory requirement from Plenoxels by a factor of 200 when using CP. Their VM factorization performed better in terms of visual quality, albeit at a memory tradeoff. The training speed was comparable to Pleoxels and much faster than the implicit NeRF models. 

\textbf{Streaming Radiance Fields} \cite{2022streaming} (October 2022) is an explicit representation method that specifically targeted NeRF training from video and improved on standard explicit methods. The authors employed model difference-based compression to reduce the memory requirement of the explicit representation. The method also uses a narrow-band tuning method and various training acceleration strategies. This method achieved a training time approximately 90 times faster than Plenoxels \cite{2021plenoxels}, with a memory requirement that is 100 to 300 times smaller.

\blu{\textbf{Dictionary Fields/Factor Fields} \cite{2023hybrid} (February 2023) proposes factoring a neural field into a spatial coefficient field times a shared basis field, using periodic coordinate warping to improve quality-per-parameter and accelerate training across image regression, signed distance fields, and radiance fields. Dictionary Fields enable multi-signal learning and few-shot or sparse-view reconstruction because the basis captures reusable structure while coefficients remain scene-specific, and note that the basis field representation choice adds extra flexibility.}

\subsubsection{Ray Transformers}
\textbf{IBRNet} \cite{2021ibrnet} (February 2021) was published in 2021 as a NeRF adjacent method for view synthesis that is widely used in benchmarks. For a target view, IBRNet selected N views from the training set whose viewing directions are most similar. A CNN was used to extract features from these images. For a single query point, for each of the $i$ input views, the known camera matrix was used to project onto the corresponding image to extract color $\V{c}_i$ and feature $\V{f}_i$. An MLP was then used to refine these features $\V{f}'_i$ to be multi-view aware and produce pooling weights $w_i$. For density prediction, these features were summed using the weights. This is done for each query point, and the results (of all query points along the ray) were concatenated together and fed into a ray Transformer \cite{vaswani2017attention}, which predicted the density. 

Compared to NeRF models, \textbf{Scene Rendering Transformer (SRT)} (November 2021) \cite{2022SRT} took a different approach to volume rendering. They used a CNN to extract feature patches from scene images which were then fed into an Encoder-Decoder Transformer \cite{vaswani2017attention} along with camera ray and viewpoint coordinates $\{\mathbf{o},\mathbf{d}\} $, which then produced the output color. The entire ray was queried at once, unlike with NeRF models. The SRT is geometry-free. and did not produce the scene's density function, nor did it rely on geometric inductive biases. 

The \textbf{NeRFormer} (September 2021) \cite{2021nerformer} is a comparable concurrent model that also uses Transformers as part of the volume rendering process. NerFormer processes ray-depth-ordered feature sequences from multiple source views using alternating pooling and ray-wise attention layers, enabling effective joint feature aggregation and ray marching. The paper also introduced the Common Objects in 3D dataset.

%% file: figures/figure_innovation_taxonomy.tex
\begin{figure*}[htbp]
 \centering
 \resizebox{\textwidth}{!}{%
\begin{forest}
  forked edges,
  for tree={
    grow=east,
    reversed=true,
    anchor=base west,
    parent anchor=east,
    child anchor=west,
    base=left,
    font=\small,
    rectangle,
    draw={hiddendraw, line width=0.6pt},
    rounded corners,align=left,
    minimum width=2.5em,
    edge={black, line width=0.55pt},
    l+=1.9mm,
    s sep=7pt,
    inner xsep=7pt,
    inner ysep=8pt,
    ver/.style={rotate=90, rectangle, draw=none, rounded corners=3mm, fill=red, text centered,  text=white, child anchor=north, parent anchor=south, anchor=center, font=\fontsize{10}{10}\selectfont,},
    level2/.style={rectangle, draw=none, fill=orange,  
    text centered, anchor=west, text=white, font=\fontsize{8}{8}\selectfont, text width = 7em},
    level3/.style={rectangle, draw=none, fill=brown,   fill opacity=0.8,text centered, anchor=west, text=white, font=\fontsize{8}{8}\selectfont, text width = 2.7cm, align=center},
    level3_2/.style={rectangle, draw=none, fill=gray,   fill opacity=0.8,text centered, anchor=west, text=white, font=\fontsize{8}{8}\selectfont, text width = 2.7cm, align=center},
    level3_1/.style={rectangle, draw=none, fill=brown,   fill opacity=0.8, text centered, anchor=west, text=white, font=\fontsize{8}{8}\selectfont, text width = 3.2cm, align=center},
    level4/.style={rectangle, draw=red, text centered, anchor=west, text=black, font=\fontsize{8}{8}\selectfont, align=center, text width = 2.9cm},
    level5/.style={rectangle, draw=red, text centered, anchor=west, text=black, font=\fontsize{8}{8}\selectfont, align=center, text width = 14.7em},
    level5_1/.style={rectangle, draw=red, text centered, anchor=west, text=black, font=\fontsize{8}{8}\selectfont, align=center, text width = 26.1em},
  },
  where level=1{text width=5em,font=\scriptsize,align=center}{},
  where level=2{text width=6em,font=\tiny,}{},
  where level=3{text width=6em,font=\tiny}{},
  where level=4{text width=5em,font=\tiny}{},
  where level=5{font=\tiny}{},
  [NeRF \cite{nerf2020_mildenhall}, ver
    [Photometric\\/Geometric Quality \\(Sec. \ref{sec:fundamentals}), level2
        [mip-NeRF \cite{2021mipnerf} based, level3
                [mip-NeRF (2021) \cite{2021mipnerf} {,}
                Ref-NeRF (2021) \cite{2022refnerf} {,} RawNeRF (2021) \cite{2022nerfinthedark}, level5_1]]
        [Depth/Point Cloud \\ Supervision, level3
                [DS-NeRF (2021)\cite{2022dense_depth_nerf} (2021){,}
                NerfingMVS (2021) \cite{2021nerfingmvs} (2021){,}\\ Urban Radiance Field (2021) \cite{2022urbannerf}{,} PointNeRF (2022) \cite{2022pointnerf}, level5_1]]]
    [Speed (Sec. \ref{sec:speed}), level2
        [Non-Baked/Hybrid, level3
            [NSVF (2020) \cite{nerf2020_NSVF}{,}
            AutoInt (2020) \cite{2021autoint} {,}
            Instant-NGP (2022) \cite{nerf2022_ngp}, level5_1]
        ]
        [Baked, level3
            [SNeRG (2020) \cite{2021sNeRG}{,} Plenoctree (2021) \cite{nerf2021_plenoctrees}{,} FastNeRF (2021) \cite{2021fastNerf}{,}\\KiloNeRF (2021) \cite{nerf2021_kilonerf}, level5_1]
        ]
        [Explicit (\ref{sec:no mlp}), level3_2
            [Plenoxels (2021) \cite{2021plenoxels}{,} DVGO (2021) \cite{2022voxeldirect}{,} TensoRF (2022) \cite{2022tensorf}, level5_1]
        ]
    ]
    [Sparse View \\ (Sec. \ref{sec:sparse}), level2
        [Cost Volume, level3
            [MVSNeRF \cite{2021mvsnerf} (2021){,} PixelNeRF (2020) \cite{nerf2021pixelnerf}{,} NeuRay (2021) \cite{2022neuralray}
            , level5_1
        ]
    ]
        [Others, level3
            [ DietNeRF (2021) \cite{2021dietNeRF}{,} DS-NeRF (2021) \cite{2022dsnerf},level5_1
            ]
        ]
    ]
    [Generative/ \\Conditional \\
    (Sec. \ref{sec:conditional}), level2
        [GAN, level3
            [GIRAFFE (2020) \cite{2021giraffe}{,} GRAF (2020) \cite{2020graf}{,} $\pi$-GAN (2020) \cite{2021pigan} {,} \\ GNeRF (2021) \cite{2021Gnerf}{,} Stylenerf (2022) \cite{2022stylenerf}{,}
            EG3D (2022) \cite{EG3D} ,level5_1
            ]
        ]
        [Diffusion, level3
            [DreamFusion (2022) \cite{2022dreamfusion}{,} Magic3D (2022) \cite{2022magic3d}{,} RealFusion (2023) \cite{2023realfusion} , level5_1
            ]
        ]
        [GLO, level3
            [NeRF-W (2020) \cite{nerf2020_nerfw}{,} Edit-NeRF (2021) \cite{2021editnerf}{,} CLIP-NeRF (2021) \cite{2022clipnerf} , level5_1
            ]
        ]
    ]
    [Composition \\(Sec. \ref{sec:composition}), level2
        [Background, level3
            [NeRF-W (2020) \cite{nerf2020_nerfw}{,} NeRF++ (2020) \cite{nerf2020nerf++}{,} GIRAFFE (2020) \cite{2021giraffe}, level5_1
            ]
        ]
        [Semantic/ \\ Object Composition, level3
            [Fig-NeRF (2021) \cite{2021fig}{,} Yang et al. (2021) \cite{2021YangObjectComposition}{,} NeSF \cite{NeSF}{,} \\
            Semantic-NeRF (2021) \cite{2021semanticnerf}{,} Panoptic Neural Fields (2022) \cite{2022panoptic}, level5_1
            ]
        ]
        [Dynamic Scene/\\
        Animation, level3
            [D-NeRF (2020) \cite{2021dnerf} HyperNeRF (2021) \cite{nerf2021_hypernerf}{,} \\
            HexPlane (2023) \cite{2023hexplane}{,} K-Planes (2023) \cite{2023kplanes}, level5_1
            ]
        ]
    ]
    [Pose Estimation \\ (Sec. \ref{sec:pose}), level2
        [SLAM, level3
            [iMAP (2021) \cite{2021imap}{,} NICE-SLAM (2021) \cite{2022niceslam}{,} NeRF-SLAM (2022) \cite{2022nerfSLAM}, level5_1
            ]
        ]
        [BA and others, level3
            [NeRF-- (2020) \cite{nerf2021_nerf--}{,} BARF (2021) \cite{2021barf}{,} \\SCNeRF (2021) \cite{nerf2021self_calibrating}{,} GARF (2022) \cite{2022garf}, level5_1
            ]
        ]
    ]
]
\end{forest}
} 
\caption{Taxonomy of selected key NeRF innovation papers pre-Gaussian Splatting. The papers are selected using a combination of citations and GitHub star rating. We note that the MLP-less explicit representation volume rendering methods are not neural fields. Nonetheless,  we decided to include them in this taxonomy tree, reflecting the state and progression of the literature in early 2023.}
\label{fig:taxonomy_innovations}
\end{figure*}

%% file: 3_5NeRF_Applications.tex
\section{Applications of NeRF and adjacent methods pre-Gaussian Splatting} \label{Applications}

This section details select works whose innovations focused on specific applications of NeRF, culminating in an organizational classification tree (Fig. \ref{fig:taxonomy_application}). The classification tree also includes certain models previously introduced in Section \ref{sec:nerf} with a strong focus on applications.

Subsection~\ref{sec:41} describes NeRF-based urban reconstruction methods developed for large-scale and unbounded outdoor environments, addressing challenges such as limited camera diversity and transient objects.
Subsection~\ref{sec:42} details techniques for reconstructing human faces, full-body avatars, and articulated objects through deformation fields, skeleton-based modeling, and generative diffusion-guided pipelines.
Subsection~\ref{sec:43} reviews NeRF adaptations for image processing tasks, including HDR rendering, denoising, deblurring, and super-resolution under varying illumination and motion conditions.
Subsection~\ref{sec:43} also contains Subsection~\ref{sec:44}, which describes semantic NeRF models that learn to synthesize semantic labels, perform segmentation, and enable scene editing through multi-view semantic fusion and 3D-aware feature distillation.
Subsection~\ref{sec:45} reviews surface reconstruction methods that replace NeRF’s implicit density representation with occupancy or signed distance fields to yield explicit and accurate 3D geometry.

A work by \textbf{Adamkiewicz et al.} (October 2021) \cite{2022navigation} focused on the localization and navigation aspect, and demonstrated a real-life application of a pretrained NeRF, in assisting the navigation of a robot through a church. The authors represented the environment with a pretrained NeRF model, with the robot itself approximated by a finite collection of points for collision checking. Since the NeRF model is pretrained, this method cannot be classified as a pose-estimation model, but instead demonstrated an interesting real-life use of NeRF.

\textbf{Dex-NeRF} \cite{2022dex} (October 2021) used NeRF's learned density to help robots grasp objects, specifically focusing on transparent objects, which were often failure cases for depth maps produced by certain RGB-D cameras such as RealSense. The paper also presented three novel datasets focused on transparent objects: one synthetic and two real-world. Dex-NeRF improved upon baseline NeRF with respect to computed depths of transparent objects by using a fixed empirical threshold for density along rays. Their NeRF model was then used to produce a depth map used by Dex-Net \cite{2017dexnet} for grasp planning. \textbf{Evo-NeRF} \cite{2022evodex} (November 2022) improved upon Dex-NeRF by reusing weights in sequential grasping, early termination, and an improved Radiance-Adjusted-Grasp Network capable of grasp planning with unreliable geometry.

In the following subsections, we classify applications of NeRF methods into urban reconstruction, human face and articulated body reconstruction, surface reconstruction, and low-level image processing, with the understanding that navigation and generative methods are covered in the previous subsection.


\input{figures/tree_application}

\subsection{Urban}\label{sec:41}

The training of an urban NeRF model poses some unique challenges. First, outdoor environments are unbounded; second, the camera poses typically lack variety; third, large-scale scenes are desired. We detail in this section how these models overcome some or all of these challenges. 

\textbf{Urban Radiance Fields} \cite{2022urbannerf} (November 2021) aimed at applying NeRF-based view synthesis and 3D reconstruction for urban environments using sparse multi-view images supplemented by LiDAR data. In addition to the standard photometric loss, they also use a LiDAR-based depth loss $L_{depth}$ and sight loss $L_{sight}$, as well as a skybox-based segmentation loss $L_{seg}$. These are given by
\begin{equation}
    L_{depth} = \mathbb{E}[(z-\hat{z}^2)],\;
\end{equation}

\begin{equation}
    L_{sight} = \mathbb{E}[\int_{t_1}^{t_2} (w(t) - \delta (z))^2 dt].
\end{equation}

\begin{equation}
    L_{seg} = \mathbb{E}[S_i(\V{r}\int_{t_1}^{t_2} (w(t) - \delta (z))^2 dt].
\end{equation}
$w(t)$ is defined as $T(t) \sigma(t)$ as defined in Eq. (\ref{eq:transmisivity}). $z$ and $\hat{z}$ are the LiDAR measured depth and estimated depth Eq. (\ref{eq:expected depth}), respectively. $\delta(z)$ is the Dirac delta function. $S_i(\V{r})=1$ if the ray goes through a sky pixel in the ith image, where sky pixels are segmented through a pretrained model \cite{2018atrous}, 0 otherwise. The depth loss forces the estimated depth $\hat{z}$ to match the LiDAR-acquired depth. The sight loss forces the radiance to be concentrated at the surface of the measured depth. The segmentation loss forces point samples along rays through to the sky pixels to have zero density. 3D reconstruction was performed by extracting point clouds from the NeRF model during volumetric rendering. A ray was cast for each pixel in the virtual camera. Then, the estimated depth was used to place the point cloud in the 3D scene. Poisson Surface Reconstruction was used to reconstruct a 3D mesh from this generated point cloud.

\textbf{Mega-NeRF} \cite{2022_meganerf} (December 2021) performed large-scale urban reconstruction from aerial drone images. Mega-NeRF used a NeRF++\cite{nerf2020nerf++} inverse sphere parameterization to separate foreground from background. However, the authors extended the method by using an ellipsoid, which better fits the aerial point of view. They incorporated the per-image appearance embedding code of NeRF-W \cite{nerf2020_nerfw} into their model as well. They partitioned the large urban scenes into cells, each one represented by its own NeRF module, and trained each module on only the images with potentially relevant pixels. For rendering, the method also cached a coarse rendering of densities and colors into an octree. 

\textbf{Block-NeRFs} \cite{2022blocknerf} (February 2022) performed city-scale NeRF-based reconstruction from 2.8 million street-level images. Such large-scale outdoor datasets posed problems, such as transient appearance and moving objects. Each individual Block-NeRF was built on mip-NeRF \cite{2021mipnerf} by using its IPE, and NeRF-W\cite{nerf2020_nerfw} by using its appearance latent code optimization. Moreover, the authors used semantic segmentation to mask out transient objects such as pedestrians and cars during NeRF training. A visibility MLP was trained in parallel, supervised using the transitivity function Eq. (\ref{eq:transmisivity}) and the density value generated by the NeRF MLP. These were used to discard low-visibility Block-NeRFs. Neighborhoods were divided into blocks, on which a Block-NeRF was trained. These blocks were assigned with an overlap, and images were sampled from overlapping Block-NeRFs and composited using inverse distance weighting after an appearance code matching optimization.

Other influential methods, such as  \textbf{S-NeRF} \cite{2021snerf} (April 2021), \textbf{BungeeNeRF}\cite{2022bungeenerf} (December 2021), also perform NeRF-based urban 3D reconstruction and view synthesis, albeit from remote sensing images.

\subsection{Image Processing}\label{sec:43}

Mildenhall et al. created \textbf{RawNeRF} (November 2021) \cite{2022nerfinthedark}, adapting Mip-NeRF \cite{2021mipnerf}, to High Dynamic Range (HDR) image view synthesis and denoising. RawNeRF is rendered in a linear color space using raw linear images as training data. This allowed for varying exposure and tone-mapping curves, essentially applying the post-processing after NeRF rendering instead of directly using post-processed images as training data. RawNeRF is supervised with variable exposure images, with the NeRF models' ``exposure" scaled by the training image's shutter speed, as well as a per-channel learned scaling factor. It achieved impressive results in night-time and low-light scene rendering and denoising. RawNeRF is particularly suited for scenes with low lighting. 

Concurrent to RawNeRF, \textbf{HDR-NeRF} (November 2021) \cite{2022hdrnerf} from Xin et al. also worked on HDR view synthesis. However, HDR-NeRF approached HDR view synthesis by using Low Dynamic Range training images with variable exposure times as opposed to the raw linear images in RawNeRF. RawNeRF modelled a HDR radiance $\V{e(\V{r})} \in [0,\infty)$ which replaced the standard $\V{c}(\V{r})$ in Eq. (\ref{eq:nerf}). HDR-NeRF was built on the baseline NeRF \cite{nerf2020_mildenhall}, using the same positional encoding and sampling strategy. The model was trained on a synthetic HDR dataset collected by the authors. HDR-NeRF strongly outperformed baseline NeRF and NeRF-W \cite{nerf2020_nerfw} on Low Dynamic Range (LDR) reconstruction, and achieved high visual assessment scores on HDR reconstruction.

NeRF-based image restoration methods address motion blur and noise by embedding degradation models directly into differentiable volumetric rendering. \textbf{DeblurNeRF} (published November 2021) \cite{2022_deblurnerf} learns a deformable sparse kernel that warps ray origins to approximate spatially varying blur, relying on an MLP to model local ray mixing so that training requires only blurry inputs and inference yields sharp views once the kernel is removed. \textbf{BAD-NeRF} (published November 2022) \cite{2023bad} instead models the physical exposure process by assigning each image a start and end pose, interpolating a continuous SE3 trajectory during the shutter interval, rendering sharp intermediate views, and averaging them to reproduce the observed blur while jointly optimizing the radiance field and camera motion. DeblurNeRF offers a compact and efficient approximation suited for moderate blur, whereas BAD-NeRF provides a physically grounded formulation that recovers sharper results and improves pose accuracy under severe motion blur at the cost of higher computational complexity. Beyond these two approaches, several later NeRF variants incorporated priors from diffusion models or robust regularization to improve deblurring or denoising under sparse views or challenging illumination.

NeRF super-resolution has trended from exploiting the internal 3D consistency and continuity of NeRF representation, branching towards leveraging generative priors to assist in learning high-resolution details. \textbf{NeRF-SR} (December 2021) \cite{2022nerfsr} attacks the problem by deterministic supersampling that enforces sub-pixel multi-view consistency and by a depth guided patch warp and refine network that reuses one high-resolution reference image, so its extra details come from propagating observed textures rather than learning a generative prior. \textbf{Super-NeRF} (April 2023) \cite{2024sr1} instead couples NeRF with a GAN based 2D super-resolution backbone and per view latent codes, so it explicitly searches the latent space of a powerful SR prior to generate plausible yet view consistent details, and it jointly optimizes a high-resolution NeRF that is supervised only through the downsampled consistency constraint, in contrast to NeRF-SR which stays closer to classical supersampling and patch based refinement. \textbf{ZS-SRT} (December 2023) \cite{2024sr2} differs from both in that it does not rely on any external SR prior or HR reference images but instead first learns a scene specific degradation model on a coarse NeRF, then trains a fine super-resolution NeRF via inverse rendering that backpropagates gradients through this degradation, together with a temporal ensemble at inference, which makes the whole pipeline fully zero shot and single scene internal and also accelerates training through an explicit coarse to fine schedule.

\textbf{NaN} (April 2022) \cite{2022nan} incorporated inter-view and spatial awareness, enhancing noise robustness, achieving state-of-the-art results in burst denoising under challenging conditions like large motion and high noise. Building on IBRNet, which generalizes to unseen scenes with minimal input, the approach avoided per-scene training.

\subsubsection{Semantic NeRF Models} \label{sec:44}
The training of the NeRF model with semantic understanding or capability of semantic view synthesis is a key area of development for NeRF research pre-Gaussian Splatting. Many of the subsequent Gaussian-Splatting-based semantic view synthesis and scene understanding models were built upon previous NeRF-based approaches.

\textbf{Semantic-NeRF} (March 2021) \cite{2021semanticnerf}  was a NeRF model capable of synthesizing semantic labels for novel views. This was accomplished using an additional direction-independent MLP (branch) that took position and density MLP features as input and produced a point-wise semantic label $\V{s}$. The semantic labels were also generated via volume rendering by

\begin{equation}
S(\V{r}) = \sum_i^N T_i \alpha_i \V{s}_i.
\end{equation}
The semantic labels were supervised using categorical cross-entropy loss. The method was able to train with sparse semantic label data (10\% labeled), and recover semantic labels from pixel-wise noise and region/instance-wise noise. The method also achieved good label super-resolution results and label propagation from sparse point-wise annotation. It can also be used for multi-view semantic fusion, outperforming non-deep learning methods. The previously introduced Fig-NeRF \cite{2021fig} also employed a similar approach.

\textbf{Panoptic NeRF} \cite{2022panoptic3d2d} (March 2022) specialized in urban environments, focusing on 3D-to-2D label propagation, a key task in extending urban autonomous driving datasets. The method used two semantic fields, one learned by a semantic head, and another determined by 3D bounding boxes. According to the authors, the rigid bounding box-based semantics forced the model to learn the correct geometry, whereas the learned semantic head improved semantic understanding. Their method was evaluated on the KITTI-360 \cite{liao2022kitti}, outperforming previous methods of semantic label transfer. 

\textbf{Panoptic Neural Fields (PNF)} \cite{2022panoptic} (May 2022) first separated the ``stuff" (as named by the authors), considered to be the background static objects, from the ``things", considered to be the moving objects in the scene. The ``stuff" was represented by a single (two in large scenes, one for foreground, one for background) radiance field MLP, which output color, density, and semantic logits, whereas each dynamic ``thing" was represented by its own radiance field inside a dynamic bounding box. The total loss function was the sum of the photometric loss function and the per-pixel cross-entropy function. The model was trained and tested on KITTI \cite{2012kitti} and KITTI 360 \cite{liao2022kitti}. In addition to novel-view synthesis and depth prediction synthesis, the model was also capable of semantic segmentation synthesis, instance segmentation synthesis, and scene editing via manipulating object-specific MLPs.

Kobayashi et al. (May 2022) \cite{2022decomposing} distilled the knowledge of an off-the-shelf 2D feature extractor into 3D feature fields, which they optimized in conjunction with in-scene radiance fields to produce a NeRF model with semantic understanding that allowed for scene editing. The distillation from the CLIP-based feature extractor allowed for zero-shot segmentation from an open set of text labels or queries.

\textbf{SS-NeRF} (June 2022) \cite{2023beyondrgb} employed an encoding function and two position decoding functions for semantics, one direction-dependent and one direction-independent, all represented by multi-layer perceptrons. The network was trained to produce a variety of scene properties, tested on the Replica dataset \cite{2019replica}: color, semantic labels, surface normal, shading, keypoints, and edges using a combination of losses including MSE for color and surface normals; MAE for shading, keypoints and edges; and cross-entropy for semantic labels. This work showed that scene property synthesis was easily achievable via volume rendering and simple NeRF training without making use of advanced neural architectures.



\subsection{Surface Reconstruction} \label{sec:45}
The scene geometry of the NeRF model is implicit and hidden inside the neural networks. However, for certain applications, more explicit representations, such as 3D mesh, are desired. For the baseline NeRF, it is possible to extract a rough geometry by evaluating and thresholding the density MLP. The methods introduced in this subsection used innovative scene representation strategies that change the fundamental behavior of the density MLP. Strictly speaking, these methods are not NeRF and are instead categorized as general neural fields. Post-Gaussian Splatting, authors tend to emphasize this difference.

\textbf{UNISURF} (April 2021) \cite{2021unisurf} reconstructed scene surfaces by replacing the alpha value $a_i$ at the i-th sample point used in the discretized volume rendering equation, given by Eq. (\ref{eq:alpha}), with a discrete occupancy function $o(\mathbf{x}) = 1$ in occupied space, and $o(\mathbf{x}) = 0$ in free space. This occupancy function was also computed by an MLP and essentially replaced the volume density. Surfaces were then retrieved via root-finding along rays. UNISURF outperformed benchmark methods, including using a density threshold in baseline NeRF models, as well as IDR \cite{2020IDR}. The occupancy MLP was used to define an explicit surface geometry for the scene. A recent presentation workshop by Tesla \cite{tesla_WAD_22} showed that the autonomous driving module's 3D understanding was driven by one such NeRF-like occupancy network.

The \textbf{Neural Surface (NeuS)} \cite{2021neus} (June 2021) model performed volume rendering like the baseline NeRF model. However, it used signed distance functions to define scene geometries. It replaced the density-outputting MLP an MLP that outputs the signed distance function value. The density $\rho(t)$, which replaced $\sigma(t)$ in the volume rendering equation Eq. (\ref{eq:ray}), was then constructed as

\begin{equation}
    \rho(t) = \max\left(\frac{-\frac{d\Phi}{dt} ( f(\mathbf{r}(t)))}
    {\Phi (f(\mathbf{r}(t)))},0\right)
\end{equation}

where $\Phi(\cdot)$ was the sigmoid function, and its derivative $\frac{d\Phi}{dt}$ was the logistic density distribution. The authors showed that their model outperformed the baseline NeRF model and provided both theoretical and experimental justifications for their method and its implementation of SDF-based scene density.
\textbf{HF-NeuS} (June 2022) \cite{2022hfneus} improved on NeuS by separating low-frequency details into a base SDF and high-frequency details into a displacement function, greatly increasing reconstruction quality. Concurrently, \textbf{Geo-NeuS} (May 2022) \cite{2022GeoNeus} introduced new multi-view constraints in the form of a multi-view geometry constraint for the SDF supervised by a sparse point cloud, and a multi-view photometric consistency constraint. \textbf{SparseNeus} (June 2022) \cite{2022sparseneus}, also concurrent, improved on NeuS by focusing on sparse-view SDF reconstruction using a geometry encoding volume with learnable image features as a hybrid representation method.

A concurrent work by \textbf{Azinovic et al.} \cite{2022neuralrgbd_surface} (April 2021) also replaced the density MLP with a truncated SDF MLP. They instead computed their pixel color as the weighted sum of sampled colors. 
\begin{equation}
    C(\V{r}) = \frac{\sum_{i=1}^{N}w_i \V{c}_i}{\sum_{i=1}^{N}w_i}.
\end{equation}
$w_i$ was given by a product of sigmoid functions:
\begin{equation}
    w_i = \Phi \left(\frac{D_i}{tr}\right) \cdot \Phi\left(-\frac{D_i}{tr}\right)
\end{equation}
where $tr$ was the truncation distance, which cut off any SDF value too far from individual surfaces. To account for possible multiple ray-surface intersections, subsequent truncation regions were assigned a weight of zero and did not contribute to the pixel color. The authors also used a per-frame appearance latent code from NeRF-W \cite{nerf2020_nerfw} to account for white-balance and exposure changes. They reconstructed the triangular mesh of the scene by using Marching Cubes \cite{1987marchingcubes} on their truncated SDF MLP and achieved clean reconstruction results on ScanNet \cite{2017scannet} and a private synthetic dataset.

\subsection{Human Faces and Body Avatars, and Articulated Objects} \label{sec:42}
A key application of NeRF models was the reconstruction of human avatars, finding applications in virtual/augmented reality, digital entertainment, and communication. Two families of NeRF models targeted these applications: those that reconstructed human (or animal) faces and those that reconstructed human/articulated bodies. The reconstruction of human faces required the NeRF model to be robust under changes of facial expression, which often manifested as topological changes. Models typically parameterized the deformation field as additional MLP(s), potentially conditioned by latent code(s), allowing for controlled deformation from a baseline human face (see subsection \ref{sec:deformation}). It is worth noting that many of the GAN-based NeRF models or NeRF models in GAN frameworks (subsection \ref{sec:gan}) were trained and tuned on datasets of human faces such as CelebA \cite{2016celeba} or FFHQ \cite{2019styleGANFFHQ}, and could arguably be placed in this section. The human body poses a different set of challenges. The NeRF model had to be robust under pose changes for articulated bodies, which were often modeled as a deformation field with a template human body model.

\label{sec:deformation}

Park et al. introduced \textbf{Nerfies} (November 2020) \cite{nerf2021_nerfies}, a NeRF model built using a deformation field which strongly improved the performance of their model in the presence of non-rigid transformations in the scene (e.g., a dynamic scene). By introducing an additional MLP that mapped input observation frame coordinates to deformed canonical coordinates and by adding elastic regularization, background regularization, and coarse-to-fine deformation regularization by adaptive masking the positional encoding, they were able to accurately reconstruct certain non-static scenes which the baseline NeRF completely failed to do. An interesting application the authors found was the creation of multi-view ``selfies" \footnote{Popular self-portraits in social media}. Concurrent to Nerfies was NerFace \cite{2021nerface} (December 2020), which also used per-frame learned latent codes, and added facial expression as a 76-dimensional coefficient of a morphable model constructed from Face2Face \cite{2016face2face}. Subsequently, Park et al. also introduced \textbf{HyperNeRF} (June 2021) \cite{nerf2021_hypernerf}, which built on Nerfies by extending the canonical space to a higher dimension and using a slicing MLP that described how to return to the 3D representation using ambient space coordinates. The canonical coordinate and ambient space coordinate were then used to condition the usual density and color MLPs of baseline NeRF models. HyperNeRF achieved great results in synthesizing views in scenes with topological changes with examples such as a human opening their mouth, or a banana being peeled.

\begin{figure}[htpb!] 
\centering
\includegraphics[width=0.49\textwidth]{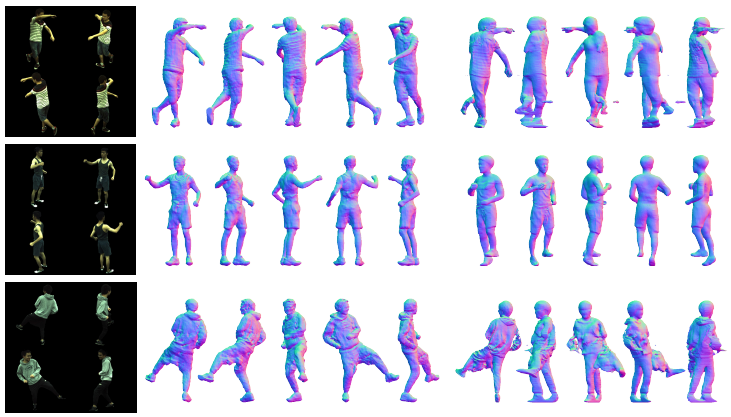}
\caption{Visualization of the 3D reconstruction results of Neural Bodies \cite{2021neuralbody} on the ZJU-Mocap dataset. Left-to-right: input views, Neural Bodies results, PIFuHD results. ©2021 IEEE} \label{fig:neuralbodies}
\end{figure}

\textbf{Neural Body} \cite{2021neuralbody} (Dec 2020) applied NeRF volume rendering to human avatars with moving poses from videos. The authors first used the input video to anchor a vertex-based deformable human body model (SMPL \cite{2015SMPL}). Onto each vertex, the authors attached a 16-dimensional latent code $\V{Z}$. Human pose parameters $\V{S}$ (initially estimated from video during training, can be input during inference) were then used to deform the human body model. The use of a baseline SMPL skeleton model with a neural deformation field became a foundational method in the neural field rendering of human avatars. The results are visualized in Figure \ref{fig:neuralbodies}.

\textbf{NELF} (July 2021) \cite{sun2021nelf} presented a neural volumetric rendering framework that modeled scene appearance using light transport vectors, enabling realistic relighting and view synthesis of human portraits from only five input images. A UNet-style CNN extracted per-view features, and an MLP regressed volume density and transport vectors, while environment maps were estimated to disentangle lighting. Trained on synthetic data and adapted to real images via a domain adaptation module, the approach achieved photo-realistic, lighting-consistent renderings, and outperformed existing methods in both quality and efficiency.

\textbf{CoNeRF} \cite{2022conerf} (December 2021) was built on HyperNeRF, but allowed for easily controllable photo editing via sliders, whose values were provided to a per-attribute Hypermap deformation field, parameterized by an MLP. This was done via sparse supervised annotation of slider values and image patch masks, with an $L2$ loss term for slider attribute value, and a cross-entropy loss for mask supervision. CoNeRF achieved good results, using sliders to adjust facial expressions in their example dataset, which could have broad commercial applications for virtual human avatars. \textbf{RigNeRF} \cite{rignerf} (June 2022) also innovated on this topic by using deformation fields MLP guided by a morphable 3D face model, creating a fully 3D face portrait with controllable pose and expression.

Standard NeRF approaches struggled with moving bodies, whereas the mesh deformation approach of Neural Body was able to interpolate between frames and between poses. A popular paradigm for animating an articulated body was established, using a baseline skeleton and equipped on top of it either an MLP-based deformation field or some other implementation of a neural field. In the following two years, this inspired a large number of works such as \textbf{A-NeRF}\cite{2021anerf} (Feb 2021), \textbf{Animatable NeRF} \cite{2021_ICCV_Animate} (May 2021) and its follow-up paper \textbf{Animatable Implicit Neural Representation} (15 March 2022) \cite{2024animatable}, \textbf{DoubleField} \cite{2022_DoubleField} (Jun 2021), \textbf{HumanNeRF} \cite{2022humannerf} (Jan 2022), \textbf{Zheng et al. \cite{2022_Zhengetal}}(March 2022), \textbf{NeuMan} \cite{2022neuman} (March 2022), \textbf{PINA} (March 2022) \cite{2022pina} \textbf{TAVA} \cite{2022tava} (June 2022), \textbf{Fast-SNARF} (November 2022) \cite{2023fastsnarf}, \textbf{ELICIT} (December 2022) \cite{2022elicit}, \textbf{X-Avatar} (March 2023) \cite{2023xavatar} which all innovated on this topic.

\textbf{PREF} \cite{2022pref} (September 2022) in particular focused on dynamics and motion in image sequences by regularizing estimated motion conditioned on latent embedding. Although PREF was trained and tested on image sequences of human avatars, it should apply to other domains. Many of the aforementioned papers, such as \textbf{NeuMan} and \textbf{TAVA}, also focus on animating the subject under novel (human subject) poses and motions. 

\textbf{LISA} (April 2022) specifically targeted the modeling of hands by approximating human hands with a collection of rigid parts. The query points were input into MLPs, which were used to predict geometry (via SDFs) and color. 

The other popular sub-area of research focused on face avatars, with constraints/requirements based on animating expressions or face topology. This research area is continuing from and improving upon the pioneering research in HyperNeRF \cite{nerf2021_hypernerf} and NeRFies \cite{nerf2021_hypernerf}. Some impactful works were \textbf{Neural Head Avatar} (December 2021) \cite{2022neuralhead}, \textbf{IMAvatar} (December 2021) \cite{2022avatar}, \textbf{INSTA} (November 2022) \cite{2023instanthead}.

In 2022, an emerging research area was diffusion-based 3D avatar model generation with text guidance powered by neural fields and NeRF. \textbf{DreamAvatar} (April 2023) \cite{2023dreamavatar}, \textbf{DreamHuman} (June 2023) \cite{2023dreamhuman}, \textbf{AvatarVerse} (August 2023) \cite{2023avatarverse} were conceptually similar, using an SMPL model as a shape prior, and used text-guided 2D image generation via diffusion to create training data in a DreamFusion-like \cite{2022dreamfusion} 3D generation pipeline combining NeRF and diffusion.

%% file: figures/tree_application.tex
\begin{figure*}[htbp]
 \centering
 \resizebox{\textwidth}{!}{%
\begin{forest}
  forked edges,
  for tree={
    grow=east,
    reversed=true,
    anchor=base west,
    parent anchor=east,
    child anchor=west,
    base=left,
    font=\small,
    rectangle,
    draw={hiddendraw, line width=0.6pt},
    rounded corners,align=left,
    minimum width=2.5em,
    edge={black, line width=0.55pt},
    l+=1.9mm,
    s sep=7pt,
    inner xsep=7pt,
    inner ysep=8pt,
    ver/.style={rotate=90, rectangle, draw=none, rounded corners=3mm, fill=red, text centered,  text=white, child anchor=north, parent anchor=south, anchor=center, font=\fontsize{10}{10}\selectfont,},
    level2/.style={rectangle, draw=none, fill=orange,  
    text centered, anchor=west, text=white, font=\fontsize{8}{8}\selectfont, text width = 7em},
    level3/.style={rectangle, draw=none, fill=brown,   fill opacity=0.8,text centered, anchor=west, text=white, font=\fontsize{8}{8}\selectfont, text width = 2.7cm, align=center},
    level3_1/.style={rectangle, draw=none, fill=brown,   fill opacity=0.8, text centered, anchor=west, text=white, font=\fontsize{8}{8}\selectfont, text width = 3.2cm, align=center},
    level4/.style={rectangle, draw=red, text centered, anchor=west, text=black, font=\fontsize{8}{8}\selectfont, align=center, text width = 2.9cm},
    level5/.style={rectangle, draw=red, text centered, anchor=west, text=black, font=\fontsize{8}{8}\selectfont, align=center, text width = 15.2em},
    level5_1/.style={rectangle, draw=red, text centered, anchor=west, text=black, font=\fontsize{8}{8}\selectfont, align=center, text width = 26.1em},
  },
  where level=1{text width=5em,font=\scriptsize,align=center}{},
  where level=2{text width=6em,font=\tiny,}{},
  where level=3{text width=6em,font=\tiny}{},
  where level=4{text width=5em,font=\tiny}{},
  where level=5{font=\tiny}{},
  [Applications, ver
    [Urban Modeling \\ (Sec \ref{sec:41}), level2
        [Street Level, level3
                [Urban Radiance Field \cite{2022urbannerf}{,}
                BlockNeRF \cite{2022blocknerf}{}, level5_1]
        ]
        [Remote/Aerial, level3
                [MegaNeRF \cite{2022_meganerf}{,}
                BungeeNeRF \cite{2022bungeenerf}{,}\\ S-NeRF \cite{2021snerf}, level5_1]
        ]
    ]
    [Image Processing \\ (Sec \ref{sec:43}), level2
        [Editing, level3
            [ClipNeRF \cite{2022clipnerf}{,} EditNeRF \cite{2021editnerf}{,}
            CodeNeRF \cite{2021codenerf}{,} Yang et al. \cite{2021YangObjectComposition}{,} \\CoNeRF \cite{2022conerf}, level5_1]
        ]
        [Semantics, level3
            [Semantic-NeRF \cite{2021semanticnerf}{,} NeSF \cite{NeSF}{,} Fig-NeRF \cite{2021fig}{,} \\
            Panoptic Neural Fields \cite{2022panoptic}, level5_1]
        ]
        [Fundamental \\ Operations, level3
            [HDR/Tone Mapping, level4
                [RawNeRF~\cite{2022nerfinthedark}{,} 
                HDR-NeRF \cite{2022hdrnerf}{}, level5]
            ]
            [Denoising/Deblurring/ \\
            Super-Resolution, level4
                [RawNeRF~\cite{2022nerfinthedark}{,} DeblurNeRF \cite{2022_deblurnerf}{,}\\ NaN \cite{2022nan}{,} NeRF-SR \cite{2022nerfsr}, level5]
            ]
        ]
    ]
    [Generative Models \\ (Sec \ref{sec:conditional}), level2
        [GAN, level3
            [GIRAFFE \cite{2021giraffe}{,} GRAF \cite{2020graf}{,} $\pi$-GAN  \cite{2021pigan} {,} \\ GNeRF  \cite{2021Gnerf} \cite{2021VaeNerf}{,} Stylenerf  \cite{2022stylenerf}{,}
            EG3D \cite{EG3D} ,level5_1
            ]
        ]
        [Diffusion, level3
            [DreamFusion  \cite{2022dreamfusion}{,} Magic3D  \cite{2022magic3d}{,} RealFusion \cite{2023realfusion}, level5_1
            ]
        ]
    ]
    [3D Reconstruction \\ (Sec \ref{sec:45}), level2
        [SDF, level3
            [NeuS \cite{2021neus}{,}
            Neural RGB-D \cite{2022neuralrgbd_surface}{,}
            Geo-NeuS \cite{2022GeoNeus}{,}
            HF-NeuS \cite{2022hfneus}, level5_1
        ]
        ]
        [Occupancy, level3
            [UNISURF \cite{2021unisurf},level5
        ]
    ]
    ]
    [Human Modeling \\ (Sec \ref{sec:42}), level2
        [Face, level3
            [Nerfies \cite{nerf2021_nerfies}{,} HyperNeRF \cite{nerf2021_hypernerf}{,} 
            RigNeRF \cite{rignerf}{,}
            EG3D \cite{EG3D}, level5_1
            ]
        ]
        [Body, level3
            [Neural Body \cite{2021neuralbody}{,} HumanNeRF \cite{2022humannerf}{,}
            Zheng et al. \cite{2022_Zhengetal}{,} \\ DoubleField \cite{2022_DoubleField}{,} LISA \cite{2022lisa}{,} Animatable NeRF \cite{2021_ICCV_Animate}{,} NeuMan \cite{2022neuman}, level5_1
            ]
        ]
    ]
    [Pose Estimation \\ (Sec. \ref{sec:pose}), level2
        [SLAM, level3
            [iMAP (2021) \cite{2021imap}{,} NICE-SLAM  \cite{2022niceslam}{,} NeRF-SLAM \cite{2022nerfSLAM}, level5_1
            ]
        ]
    ]
]
\end{forest}
} 
\caption{Taxonomy of selected key NeRF applications papers pre-Gaussian Splatting. Papers are classified based on application and selected using citation numbers and GitHub star ratings.}
\label{fig:taxonomy_application}
\end{figure*}

%% file: 4PostGS.tex
\section{Post-Gaussian Splatting Neural Rendering and NeRF} \label{sec:postgs}
3D Gaussian splatting \cite{2023gaussian_splatting} is a method for 3D scene representation and novel view synthesis that represents a scene using a set of anisotropic 3D Gaussians. Each Gaussian encodes position, scale, orientation, opacity, and color, allowing the scene to be rendered through a fast differentiable splatting process that projects and blends these primitives in screen space. Gaussian Splatting methods are typically much faster and produce slightly better quality images, but require more memory and storage space. 

Gaussian splatting-based methods \cite{2023gaussian_splatting} have overtaken NeRF and NeRF-adjacent Neural Rendering methods for novel-view synthesis and adjacent tasks. The shift in research momentum was so severe that implicit and hybrid neural field methods distanced themselves from the ``NeRF" keyword. Nonetheless, these methods remained popular in certain applications where implicit neural field-based representations are desirable. In this section, we detail the relevant implicit and hybrid neural field methods and NeRF methods. These works are organized into the following subsections. Subsection~\ref{sub:1} presents improvements in differentiable volume rendering with implicit and hybrid neural field representations, highlighting advances that improve rendering efficiency, spectral generalization, and view-dependent effects. \blu{Readers are referred to \cite{20243dgsSurvey} for an introduction to Gaussian Splatting.} 

Subsection~\ref{sub:2} describes developments in 3D scene representation, including semantic integration, vision-language grounding, and generative priors that enhance context-aware reconstruction and scene understanding.
Subsection~\ref{sub:3} details diffusion-based methods combined with neural fields for 3D generation, scene editing, and image restoration, encompassing applications in inpainting, relighting, and resolution enhancement.
Subsection~\ref{sub:4} summarizes SLAM frameworks leveraging implicit and hybrid neural fields for efficient and consistent mapping, localization, and multi-agent collaboration across diverse environments.
Subsection~\ref{sub:5} reviews human avatar reconstruction with implicit and hybrid neural fields, covering both facial and full-body modeling with photorealistic rendering and dynamic animation.

\subsection{Improvements in Differentiable Volume Rendering with Implicit/Hybrid Neural Field Representation} \label{sub:1}
\textbf{NeuRBF} (September 2023) \cite{neurbf} is a hybrid neural field model that enhances representation accuracy and compactness by combining adaptive radial basis functions (RBFs) with grid-based RBF interpolation. It generalizes earlier feature grid-based neural field methods, introducing multi-frequency sinusoidal composition to extend the frequencies encoded by each basis function. These features are then decoded by an MLP for volume rendering with SDF reconstruction capabilities. NeuRBF achieves state-of-the-art performance across 2D image fitting, 3D signed distance field reconstruction, and neural radiance field synthesis.

\textbf{PIE‑NeRF} (November 2023) \cite{2024pienerf} is a novel pipeline that integrates physics-based elastodynamic simulation with a neural radiance field representation. Instead of converting the learned implicit surface into a mesh or voxel grid, the method uses a meshless discretisation based on quadratic generalized moving least squares on top of the density field of a NeRF. This enables large deformations and nonlinear hyperelastic material simulation directly on the implicit representation and supports interactive manipulation of real-world captured scenes with complex geometries.

\textbf{FastSR-NeRF} (December 2023) \cite{2024fastsr} introduces a simple super-resolution/upsampling CNN into the NeRF pipeline. The method trains a small, fast, and efficient NeRF model to produce low-resolution 3D consistent features and uses a fast SR model to upscale these features, significantly reducing the computational cost of volume rendering. Unlike prior NeRF+SR methods that rely on complex training procedures, distillation, or high-resolution reference images, FastSR-NeRF requires no architectural changes or heavy computing. It introduces a novel augmentation technique called random patch sampling, which improves SR performance by increasing patch diversity. The method is especially suitable for consumer-grade hardware, making neural rendering more accessible.

\textbf{Viewing Direction Gaussian Splatting (VDGS)} (December 2023) \cite{2025vdgs} is a hybrid approach that combines the fast, efficient rendering of Gaussian Splatting with the view-dependent modeling capabilities of NeRF. VDGS uses the 3D Gaussian Splatting representation of geometry and a NeRF-based encoding of color and opacity. VDGS inherits Gaussian Splatting's real-time inference performance while significantly reducing view-dependent artifacts. 

\textbf{MulFAGrid} (May 2024) \cite{2024mulfa} is a general-purpose grid-based neural field model that integrates multiplicative filters with Fourier features. Guided by a new Grid Tangent Kernel (GTK) theory, the method emphasizes spectral efficiency in high-frequency regimes, offering improved generalization and learning capacity over prior models such as InstantNGP \cite{nerf2022_ngp} and NeuRBF \cite{neurbf}. MulFAGrid supports both regular and irregular grids and is trained via a joint optimization of grid and kernel features. Results across 2D image fitting, 3D signed distance field reconstruction, and novel view synthesis show excellent performance, with strong results in NeRF-based benchmarks. While slower than real-time renderers like 3DGS, MulFAGrid offers a robust, flexible alternative for neural field representation.

\textbf{NU-NeRF} (November 2024) \cite{2024nunerf} proposes a method to reconstruct nested transparent objects from standard multiview images. It uses a two-stage process: in the first stage, it separates surface color into reflection and refraction components and fits an outer surface geometry via a neural signed distance field plus a refraction MLP; in the second stage, it explicitly ray-traces light through the outer surface and reconstructs inner surfaces via another signed-distance-based model with a transparent-interface formulation. The result enables accurate nested transparent-object geometry and appearance recovery under more realistic capture conditions.
 
\textbf{VD‑NeRF} (January 2025) \cite{2025vdnerf} proposes a neural radiance field architecture that decouples view-independent appearance (albedo and diffuse lighting) from view-dependent effects (specular highlights and reflections) while explicitly modelling visibility and shadowing to support consistent editing and high-frequency relighting. Its key innovations include a visibility-aware mask that guides separation of static and dynamic lighting contributions, a two-branch network (one for base appearance, one for view-dependent residuals), and a relighting pipeline that allows for scene re-illumination under new light directions thanks to the disentangled components.

\subsection{Developments in 3D Scene Representation} \label{sub:2}
\subsubsection{3D Scene Understanding, Semantics, and \blu{Dynamics}}
\textbf{GP-NeRF} (November 2023) \cite{2024gpnerf} is a unified framework that integrates NeRF with 2D semantic segmentation modules to enable context-aware 3D scene understanding. Unlike prior methods that treat semantic labels and radiance fields independently, GP-NeRF jointly learns radiance and semantic embedding fields using a Field Aggregation Transformer and a Ray Aggregation Transformer. This architecture allows joint rendering and optimization of both fields in novel views.

\blu{\textbf{FRNeRF} (March 2024) \cite{2025frnerfdynamic} tackles monocular dynamic space-time view synthesis by augmenting an NSFF-style dynamic NeRF with a 2D–3D fusion regularization field that couples a depth-free 2D feature flow with the 3D scene flow, which targets pixel misalignment and depth-induced artifacts that especially hurt fast motion. It also injects image-prior features from a masked autoencoder into hierarchical sampling, so the radiance field gets richer supervision on under-observed dynamic pixels, and it can render sharper moving regions. A ViT-based \cite{vit} global alignment loss further regularizes the rendered output at the semantic level, which stabilizes training under residual flow errors.}

\textbf{DP-RECON} (March 2025) \cite{2025dprecon} presents a decompositional 3D reconstruction method that integrates a generative diffusion prior with neural implicit representations. Given posed multiview images, the method reconstructs individual objects and backgrounds while optimizing geometry and appearance using Score Distillation Sampling (SDS) from a pretrained Stable Diffusion model. To address conflicts between the generative prior and observed data, a novel visibility-guided optimization is introduced. This visibility map, learned via a differentiable grid based on volume transmittance, modulates SDS and reconstruction loss per pixel. The method achieves high-fidelity reconstructions, especially in occluded regions, outperforming baselines even with significantly fewer input views. Additionally, it supports detailed scene editing, stylization, and outputs decomposed meshes with UV maps.

\begin{figure}[htpb] 
\centering
\includegraphics[width=0.45\textwidth]{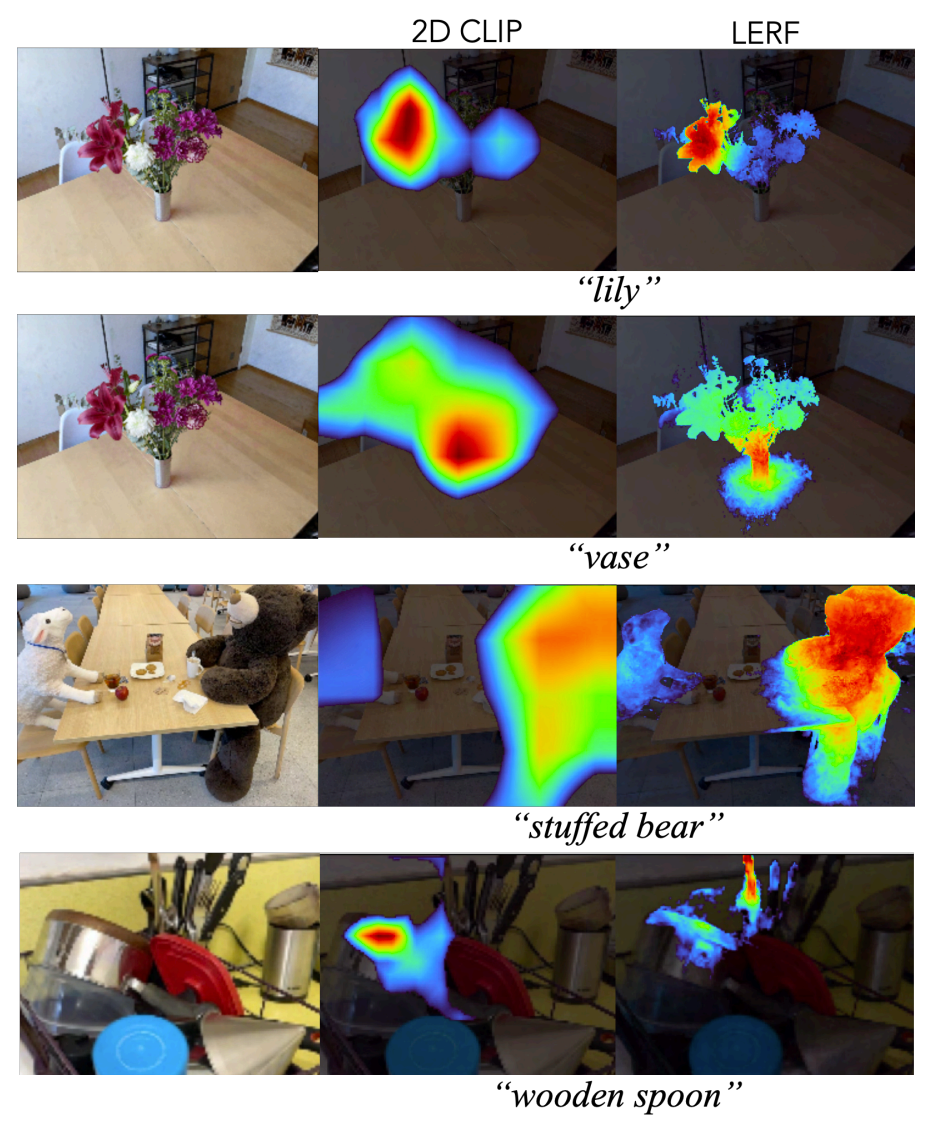}
\caption{Visualization of LERF \cite{2023lerf} embedding vision-language similarity score heatmap for different text prompts. Left: image, middle: 2D CLIP visualization (interpolated over patchwise CLIP embeddings), right: LERF visualization. ©2023 IEEE} \label{fig:LERF}
\end{figure}
\subsubsection{Language and Grounding for NeRF and Adjacent Neural Rendering Methods}

Although not strictly speaking a post-GS method, this influential paper is included in this section due to its impact on the novel and mostly post-GS research area of language grounding for NeRF representations. \textbf{Language Embedded Radiance Fields (LERF)} (May 2023) \cite{2023lerf} is a method for integrating natural language understanding directly into NeRFs by embedding CLIP features into the 3D radiance field.  LERF constructs the language field using a multi-scale feature pyramid from training views, associating each 3D location with scale-aware language semantics. To enhance semantic stability and structure, the framework also incorporates self-supervised DINO features through a shared bottleneck. The result is a model that produces 3D-consistent relevancy maps in response to natural language queries, outperforming 2D-based open-vocabulary detectors projected into 3D. LERF enables real-time, semantically aware 3D interaction, supporting use cases in robotics, scene understanding, and vision-language grounding.

\textbf{OV-NeRF} (February 2024) \cite{2024ovnerf} is a NeRF model which performs open-vocabulary 3D semantic segmentation. The method enhances single-view semantic precision using Region Semantic Ranking (RSR), which leverages region-level cues from SAM \cite{SAM} to improve boundary quality in semantic maps. To address semantic inconsistency across views, OV-NeRF introduces Cross-view Self-enhancement (CSE), which exploits NeRF's 3D consistency to refine relevancy maps and generate novel semantic views for additional supervision. These combined strategies reduce CLIP-induced ambiguities and improve multi-view coherence. Experiments on Replica and ScanNet show substantial improvements in mIoU over prior methods, demonstrating the effectiveness and generalizability of OV-NeRF in open-vocabulary 3D scene segmentation.

\textbf{Hierarchical Neural Radiance (HNR)} (April 2024) \cite{2024hnr} enhances vision-and-language navigation (VLN) by predicting robust, multi-level semantic features of future candidate environments. Leveraging CLIP-based vision-language embeddings, the model encodes 3D-aware language-aligned visual features into a hierarchical feature cloud and uses volume rendering to infer semantic context for unseen or occluded regions. This hierarchical encoding improves prediction quality and spatial understanding compared to prior 2D generation approaches. Integrated into a lookahead VLN framework comprised of a cross-modal graph encoding transformer for path planning through a future path tree. The entire framework allows for language-based path planning using a Neural Field-based 3D vision system.

\textbf{Large Language and NeRF Assistant (LLaNA)} (June 2024) \cite{2024llana} is a multimodal language model with NeRF integration. The NeRF MLP weights are embedded into the latent space of a pre-trained language model using an encoder. This approach bypasses the need to render images or extract geometry, preserving the NeRF representation. The authors also present a new NeRF–language dataset derived from ShapeNet for NeRF-based question-answering (QA) tasks. The authors also introduce large-scale training in their subsequent work \textbf{Scaling-LLaNA} (April 2025) \cite{scalellana} introducing a new large-scale NeRF-Language dataset, and additional analysis on LLM size.

\subsection{Diffusion and Neural Fields} \label{sub:3}
\subsubsection{Diffusion for 3D Generation and Editing}

 Shum et al. (September 2023) \cite{2024shum} present a language-driven 3D scene editing using text-to-image diffusion models integrated with NeRFs. The method enables object insertion and removal by synthesizing multi-view images that incorporate both the target object and background guided by a text prompt. These images are used to iteratively refine the NeRF through a pose-conditioned dataset update strategy, which gradually integrates new views to maintain consistency and stabilize training. Unlike prior approaches that rely on explicit geometry, depth, or masks, this method requires only rough user input via 3D bounding boxes. The authors demonstrate the system's ability to perform high-quality, view-consistent edits with minimal manual input and validate its effectiveness through extensive experiments, showcasing state-of-the-art results in NeRF-based scene manipulation.
 
\textbf{ReconFusion} (December 2023)\cite{2023recon} uses 2D diffusion-based priors to enhance NeRF quality, especially under sparse view conditions. A multiview-conditioned diffusion model, finetuned from a pretrained latent diffusion backbone, is trained on both real and synthetic datasets to synthesize novel views. This model serves as a regularizer within the NeRF training loop via a score distillation-like approach. The method improves reconstruction fidelity across diverse settings—mitigating artifacts such as floaters and fog in dense captures, and enabling plausible geometry in limited-view scenarios. The approach offers a general, effective prior for robust NeRF optimization.

\textbf{Comps4D} (March 2024) \cite{2024comps4d} introduces a framework for generating compositional 4D scenes (i.e. animated 3D scenes). The method moves beyond previous object-centric approaches. It decouples the process into two main stages: (1) scene decomposition to create static 3D assets and (2) motion generation guided by large language models (LLMs). The static objects are generated using a NeRF representation. The LLM plans global trajectories based on textual input, while local deformations are learned through a deformable 3D Gaussian representation. This setup allows flexible rendering and robust motion learning, even with occlusions. A compositional score distillation mechanism optimizes object dynamics. Results show superior visual fidelity, realistic motion, and coherent object interactions compared to existing methods.

\textbf{LN3Diff} (March 2024) \cite{2024ln3diff} introduces a latent-space 3D diffusion framework for conditional 3D generation. The pipeline employs a variational autoencoder to map input images into a compact, 3D-aware latent space, which is decoded into triplane representation via a transformer-based architecture. Training leverages differentiable rendering with multi-view or adversarial supervision, requiring as few as two views per scene. A convolutional tokenizer and transformer layers enable structured attention across 3D tokens, promoting coherent geometry. The latent representation supports fast amortized inference and scalable diffusion learning. LN3Diff achieves state-of-the-art performance on ShapeNet, FFHQ, and Objaverse for both 3D reconstruction and generation, outperforming existing GAN and diffusion-based baselines while offering up to 3× faster inference. 

\subsubsection{Diffusion Aiding Image Processing}

\textbf{Inpaint4DNeRF} (December 2023) \cite{2023Inpaint4DNeRF} introduces a text-guided generative NeRF inpainting method using diffusion models, with natural extension to 4D dynamic scenes. Given a user-specified foreground mask and text prompt, the method inpaints select seed views using Stable Diffusion, then infers coarse geometry from these views. The remaining views are refined with diffusion-based inpainting guided by the seed images and their geometry, ensuring multiview consistency. 

\textbf{DiSR-NeRF} (April 2024) \cite{2024disrnerf} addresses the challenge of generating high-resolution, view-consistent NeRFs from only low-resolution (LR) multi-view images, since using LR images is a common practice in NeRF training due to computational cost. Naive 2D super-resolution leads to inconsistent details across views. To fix this, DiSR-NeRF introduces two novel components. First, Iterative 3D Synchronization (I3DS) alternates between 2D diffusion-based super-resolution and NeRF training, progressively aligning details in 3D space. Second,  the Renoised Score Distillation (RSD) is introduced in this paper. It refines the diffusion process by optimizing over intermediate denoised latents to produce sharper and more consistent results. Without needing high-resolution training data, DiSR-NeRF outperforms existing methods in generating high-fidelity, super-resolved NeRFs.

 \textbf{MVIP-NeRF} (May 2024) \cite{2024mvipnerf} introduces a diffusion-based method for multiview-consistent inpainting of Neural Radiance Fields. Unlike prior approaches that rely on independent 2D inpainting per view—often resulting in inconsistencies and poor geometry, MVIP-NeRF jointly optimizes across views to ensure consistency. It employs Score Distillation Sampling (SDS) with a text-conditioned diffusion model to guide inpainting in masked regions, alongside RGB reconstruction in visible areas. To enforce geometric consistency, the method also distills normal maps. A new multi-view SDS formulation further enhances consistency under large view changes. MVIP-NeRF achieves state-of-the-art results for NeRF inpainting.

\textbf{Neural Gaffer} (June 2024) \cite{2023gaffer} presents a category-agnostic, single-view relighting framework based on a 2D diffusion model. Unlike previous models limited to specific object classes, it generalizes across arbitrary categories and lighting environments using HDR environment maps. Trained on a synthetic dataset with physically-based materials and HDR lighting, the model captures rich lighting priors, enabling accurate and high-quality relighting from a single image. Neural Gaffer outperforms existing approaches on both synthetic and real data, integrates with image editing tasks, and extends to 3D relighting via NeRFs. It establishes a versatile diffusion-based prior for relighting in both 2D and 3D domains.

\begin{figure}[htpb] 
\centering
\includegraphics[width=0.45\textwidth]{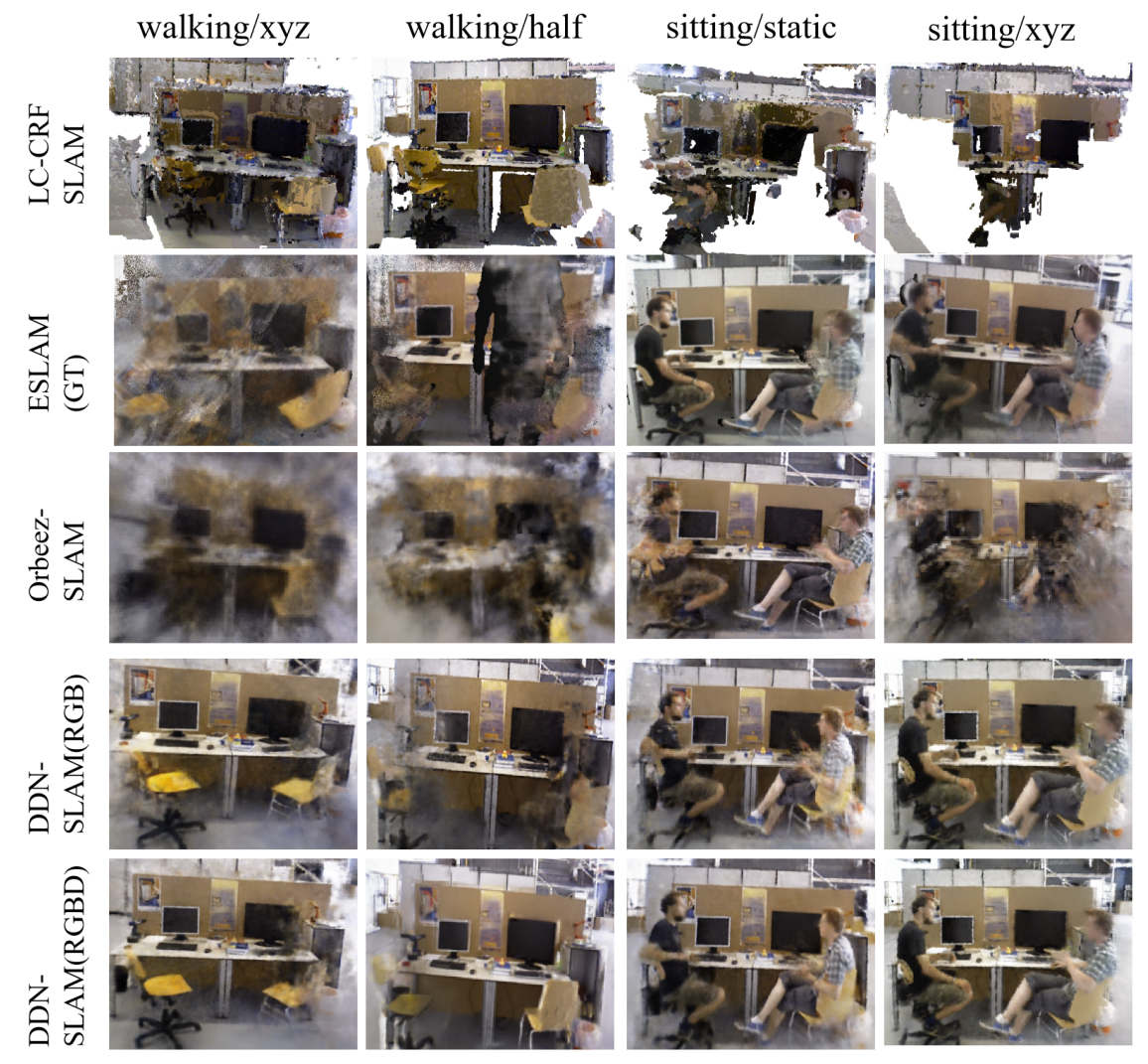}
\caption{Visualization of DDN-SLAM \cite{2024ddnslam} results on the TUM RGB-D dataset with both dynamic and static sequences, and comparisons against select benchmark SLAM methods. ©2025 IEEE} \label{fig:ddnslam}
\end{figure}
\subsection{SLAM with Implicit and Hybrid Neural Fields} \label{sub:4}

\textbf{CP-SLAM} (November 2023) \cite{2023cpslam} is a neural point-based (hybrid neural field) SLAM system that enables multi-agent cooperative localization and mapping, while supporting loop closure for individual agents. It introduces a new keyframe-associated neural point representation inspired by Point-NeRF, allowing per-point features to be easily adjusted during pose graph optimization. To ensure cross-agent consistency, CP-SLAM employs a two-stage distributed-to-centralized training scheme: initial decoders are trained separately for each agent, then fused and fine-tuned jointly. The system integrates odometry, loop detection, sub-map fusion, and global refinement into a unified framework.

\textbf{SNI-SLAM} (November 2023) \cite{2024snislam} is a dense NeRF-based RGB-D SLAM system designed for accurate real-time 3D semantic mapping. It addresses two core challenges in semantic SLAM: (1) the interdependence of appearance, geometry, and semantics, and (2) the mutual inconsistency of multi-view appearance and semantic optimization. To tackle these, SNI-SLAM introduces a hierarchical semantic encoding and cross-attention mechanism that enables mutual reinforcement between modalities. It further proposes a novel one-way decoder design to enhance inter-modal information flow without reverse interference. 

\textbf{DNS-SLAM} (November 2023) \cite{2024dnsslam} is a dense semantic-aware SLAM framework that builds on class-wise scene decomposition with a hybrid point-based Neural Field mapping module. DNS-SLAM introduces a multi-class neural scene representation that explicitly links object classes to camera poses. It leverages 2D semantic priors and multi-view image features to strengthen pose estimation through back-projected geometric constraints. A lightweight coarse model, trained via self-supervision, accelerates tracking. To further refine geometry, DNS-SLAM supervises occupancy with Gaussian-distributed priors. Visualizations of the results are shown in Fig. \ref{fig:ddnslam}. 

\textbf{Neural Graph Mapping} (December 2023) \cite{2025neuralgraphslam} introduces a dynamic multi-field scene representation composed of small, lightweight neural fields anchored to keyframes in a pose graph. These fields deform with updated poses during loop closure, enabling consistent volumetric mapping without costly reintegration or fixed scene boundaries. The proposed RGB-D SLAM framework merges accurate sparse visual tracking with dense neural mapping, achieving robust performance across diverse scenes.

\textbf{DDN-SLAM} (January 2024) \cite{2024ddnslam} integrates semantic priors with NeRF-based representation to distinguish between dynamic and static objects. The tracking and NeRF-based mapping are separated into four threads. The segmentation thread identifies and suppresses dynamic feature points and regions. The tracking thread extracts features, filters them via semantic and geometric cues, computes static optical flow, and produces camera poses and keyframes. The mapping thread integrates input sparse point cloud to guide the NeRF-based ray sampling and uses dynamic-aware masks to drive keyframe selection and volume rendering, preserving static surface geometry. The loop detection thread detects revisited areas and performs global bundle adjustment, enhancing long-range consistency. 

\textbf{PIN-SLAM} (January 2024) \cite{2024pinslam} globally consistent SLAM system using a point-based implicit neural (PIN) representation. It replaces grid structures with neural feature points, offering spatial flexibility and elastic correction during loop closure. Mapping alternates with odometry. The mapping step is based on a hybrid neural SDF representation with explicit neural points and implicit MLP decoders. Odometry is performed correspondence-free via second-order scan-to-map optimization. A sliding window replay buffer ensures stable incremental updates. Loop closures trigger pose graph optimization and elastic deformation of neural points, enabling consistent large-scale mapping. 

\textbf{KN-SLAM} (March 2024) \cite{2024knslam} integrates local feature correspondences for coarse pose initialization for the NeRF-based mapping module, jointly optimizing the photometric loss and feature reprojection loss. Global image features and local matches are used for explicit loop detection, followed by pose graph optimization and global refinement of the neural map to ensure consistency. 

\textbf{SLAIM} (April 2024) \cite{2024slaim}  introduces a coarse-to-fine tracking pipeline and improves photometric bundle adjustment through a Gaussian-filtered image signal, enhancing convergence in image alignment. It maintains NeRF’s original volume density formulation while introducing a KL regularization over the ray termination distribution. It addresses the challenge of high-frequency renderings in NeRF hindering image alignment.

\textbf{HERO-SLAM} (July 2024) \cite{2024heroslam} employs a novel multiscale patch-based loss that aligns feature points, maps, and RGB-D pixels through warpings. An INGP \cite{nerf2022_ngp}-like multi-resolution hybrid feature grid+MLP representation is used for neural SDF learning. Extensive evaluations on standard benchmarks show superior performance and robustness over prior implicit field-based SLAM methods, especially under challenging conditions.

\textbf{MNE-SLAM} (June 2025) \cite{2025mneslam} is the first fully distributed multi-agent neural SLAM framework that enables accurate collaborative mapping and robust camera tracking without centralized training or raw data exchange. The system uses a triplane+MLP hybrid neural field representation as a mapping module. It introduces an intra-to-inter loop closure strategy to reduce pose drift and align submaps across agents through peer-to-peer feature sharing and global consistency loss. To support benchmarking, the authors created the INS dataset, a real-world dataset with high-precision, time-continuous trajectories and 3D mesh ground truth, suitable for evaluating various neural SLAM systems under realistic conditions.

\blu{\textbf{GPS-SLAM} (September 2025) \cite{GPSSLAM} is an RGB-D SLAM method that represents a scene with a fast, stable geometric backbone using a truncated SDF volume and then adds a thin layer of 3D Gaussians near the surface to model high-frequency appearance residuals that the SDF color cannot capture. At render time, it raycasts the SDF for depth plus a base color, then splats only the surface-near Gaussians with SDF-guided depth culling and order-independent blending, and mixes the two images, which avoids expensive NeRF-style volumetric sampling and also removes the Gaussian depth sorting step. Empirically, this method achieves real-time to high-frame-rate operation while maintaining or improving reconstruction quality relative to Gaussian-only RGB-D SLAM baselines.}

\subsection{Human avatars with Implicit and Hybrid Neural Fields} \label{sub:5}

\subsubsection{Face}
\textbf{HQ3D} (March 2023) \cite{2024hq3d} introduces a method for generating highly photorealistic facial avatars using a voxelized feature grid with multiresolution hash encoding with decoding MLP in a hybrid neural implicit field. Trained on multi-view video data, the model operates with only monocular RGB input at test time and requires no mesh templates or space pruning. A novel canonical space, conditioned on video-extracted features, is regularized via an optical flow loss for artifact-free, temporally coherent reconstructions. The approach supports novel views and expressions, renders at 2K resolution, trains 4–5× faster than prior work, and runs in real-time. A new 4K multi-view dataset of 16 identities is also introduced. 

Qin et al. (October 2023) \cite{2024high} introduce a 3D head avatar framework that overcomes the limitations of global expression conditioning in NeRFs by proposing Spatially-Varying Expression (SVE). Unlike prior methods that use uniform global expression codes across 3D space, SVE integrates both spatial and expression features to enable fine-grained control over facial geometry and rendering. A generation network produces SVE by combining 3DMM expression parameters with position-specific features. A coarse-to-fine training strategy further refines geometry and rendering through initialization and adaptive sampling. The resulting method captures intricate details like wrinkles and eye motion with significantly higher fidelity than global-expression-based NeRFs.


\textbf{BakedAvatar} (November 2023) \cite{2023bakedavatar} proposes a novel representation for real-time 4D head avatar rendering that targets photorealism and efficiency on commodity devices. Unlike traditional mesh or NeRF-based methods, which either struggle with fine details like hair or demand heavy sampling, BakedAvatar introduces a learned manifold closely aligned with the head surface. From this, layered mesh proxies are extracted to approximate volumetric rendering while enabling fast rasterization.

\textbf{Bai et al.} (April 2024) \cite{2024efficient} propose a 3D neural avatar system that achieves real-time rendering with high fidelity and fine-grained control through mesh-anchored hash table blendshapes. Each vertex of a 3DMM mesh is linked to a local hash table, allowing for expression-dependent embeddings and localized facial deformations. These local blendshapes are combined using per-vertex weights predicted in UV space from driving signals. A hash encoding with decoding MLP in a hybrid neural implicit field is used to predict color and density from 3D queries using volume rendering.

\textbf{LightAvatar} (September 2024) \cite{2025lightavatar} a neural light field (NeLF)-based \cite{sun2021nelf} head avatar model that eliminates reliance on explicit meshes or volume rendering for a streamlined, efficient pipeline. A pretrained avatar model supervises LightAvatar via distillation. To avoid performance limitations from teacher supervision, training combines both pseudo and real data. However, since 3DMM fitting is imperfect on real data, a warping field network is introduced to correct fitting errors and enhance quality. The rendering is done in low resolution, and a super-resolution module is used to generate high-resolution images. 

\textbf{NeRFFaceShop} (October 2025) \cite{2025nerfface} learns a 3D-aware generative model that produces animatable and relightable human heads by training on large-scale in-the-wild videos and by deforming feature maps rather than 3D points so lighting and expression changes stay consistent during motion. It builds on a tri-plane generator, introduces a unified animation and lighting representation, extracts expression coefficients from video to learn a detailed animation space, and enables smooth, controllable head motion and relighting across humans and even nonhuman domains.

\subsubsection{Body}

Xu et al. (August 2023) \cite{2024relightable} presents a method for creating relightable and animatable human avatars from sparse or monocular video. The avatar is modeled as MLPs that predict material properties (light visibility, albedo, roughness) and geometric properties (SDF and surface normal) in canonical space, transformed into world space via a neural deformation field. A hierarchical distance query algorithm blends world-space KNN and canonical SDF distances, enabling accurate pixel-surface intersection via sphere tracing and improving rendering under arbitrary poses. The method also extends distance field soft shadow (DFSS) computation to deformed SDFs, allowing efficient soft shadow rendering. 

\textbf{NECA} (March 2024) \cite{2024neca} is a customizable neural avatar framework enabling photorealistic rendering under arbitrary pose, view, and lighting, while supporting fine-grained editing of shape, texture, and shadow. NECA learns human representation jointly in a canonical space and a surface-based UV-tangent space to capture both shared structure and high-frequency pose-dependent detail. Geometry, albedo, and shadow are predicted via separate MLPs, with optimized environmental lighting. Trained via self-supervision using photometric and normal constraints, the framework is built upon an SMPL model and uses attribute-based neural fields (MLPs) for SDF, albedo, and shadow.

\textbf{MeshAvatar} (July 2024) \cite{2024meshavatar} introduces a hybrid representation for triangular human avatars that enables end-to-end learning from multi-view videos by combining explicit mesh geometry with neural signed distance and material fields. The system leverages differentiable marching tetrahedra (DMTet) to bridge mesh and implicit components, allowing compatibility with traditional rendering pipelines and hardware-accelerated ray tracing. To enhance surface reconstruction and relighting, the method integrates shadow-aware physics-based rendering (PBR), pose-driven 2D neural encoders for high-frequency detail, and stereo-estimated normal maps for weak supervision. This design achieves high-quality dynamic geometry and appearance without requiring surface tracking or pre-defined templates.

Huang et al. (October 2024) \cite{2024efficient2} introduce a fast and smooth Dynamic Human NeRF model that reconstructs animatable human avatars from monocular videos. It combines HuMoR \cite{rempe2021humor} for temporally coherent pose estimation, Instant-NGP \cite{nerf2022_ngp} for accelerated canonical shape learning, and Fast-SNARF \cite{2023fastsnarf} for efficient deformation into pose space. To overcome the inefficiency of traditional volume rendering in dynamic settings, the method proposes a posture-sensitive space reduction and dynamic occupancy grid for skipping empty regions during rendering. This hybrid design significantly improves reconstruction quality and speed.

%% file: 5.Discussion_and_conclusion.tex
\section{Discussion} \label{sec:discussion}
\subsection{NeRF vs. Gaussian Splatting}

\begin{table}[h!]
\centering
\caption{Comparison of the key differences between standard NeRF and Gaussian Splatting models.}
\begin{tabular*}{\columnwidth}{@{\extracolsep{\fill}} l|ll}
\hline
\textbf{Aspect} & \textbf{NeRF} & \textbf{3DGS} \\
\hline
3D Representation & Implicit  & Explicit  \\
3D Form & Neural network fields & Gaussian points \\
Rendering & Volume rendering & Rasterization \\
Training Speed & Slow & Fast \\
Rendering Speed & Slow & Real-time \\
Memory Usage & Medium & High \\
Storage Volume & Low & High \\
View Synthesis Quality & High & Very high \\
\hline
\end{tabular*}
\label{tab:nerf_vs_3dgs}
\end{table}

NeRF and Gaussian Splatting are both novel view synthesis methods. They differ in \textbf{representation}: NeRF and adjacent neural field rendering methods use implicit or hybrid neural fields to represent the 3D scene, whereas Gaussian Splatting methods use an explicit 3D point cloud-like representation of the scene. They differ in \textbf{rendering paradigm}: NeRF and the adjacent neural field rendering methods presented in this survey use ray-tracing-like differentiable volume rendering, sampling the neural density and color field along virtual camera rays, whereas Gaussian Splatting methods use differentiable rasterization based on 2D projection of elliptical 3D Gaussian primitives (and do not explicitly sample color values along camera rays). As such, NeRF-like methods are typically more memory and storage-efficient. However, NeRF-like methods are typically much slower than Gaussian Splatting methods and often have slightly lower view synthesis quality. 

Many Gaussian Splatting methods (2023-2025) were directly adapted or drew heavy inspiration from NeRF research from the 2020-2022 era. Despite the momentum of novel view synthesis research shifting towards Gaussian Splatting in recent years, NeRF and neural field-based approaches still have certain advantages. In terms of the technical aspects, as previously mentioned, implicit and hybrid representations, such as neural fields, trade speed for memory and storage efficiency compared to explicit representations such as Gaussian Splatting. The implementation of the ``splatting"-based rasterization in Gaussian Splatting is also faster than the volume rendering approach of NeRF and neural field methods, without sacrificing view synthesis quality. However, the volumetric rendering approach is more suited for volumetric scene elements such as dust or fog: these scene elements result in floaters in a standard Gaussian Splatting approach. And finally, neural field approaches are more suited for certain computations. Neural fields can be queried on a continuum of 3D coordinates, and are well-suited for representing spatially distributed properties. This contrasts with the discrete 3D point cloud-like representation, which must be further engineered to represent spatially distributed properties.

Because of the faster training and inference time and higher view synthesis quality, Gaussian Splatting methods have largely overtaken NeRF-adjacent methods for novel view synthesis and adjacent research areas, including 3D model generation, view synthesis with scene semantics, and 3D scene representation-reconstruction-editing. This is evidenced by the much lower number and impact of implicit and hybrid neural field publications in these research areas in the post-Gaussian Splatting era.

Hybrid methods that fuse explicit Gaussian splatting and neural-field representations aim to leverage the best of both worlds: the speed and rasterization-friendliness of Gaussian splats with the flexibility, view-dependency, and expressivity of neural fields. Explicit splats provide fast rendering by representing a scene with optimized Gaussians, while neural fields bring in continuous modeling of geometry, lighting, and appearance, as well as more complex spatio-temporal mappings. Together, they address the limitations each method has in isolation.

In practice, methods that combine both Gaussian Splatting and neural fields adopt architectures where Gaussians or local feature points handle detailed local primitives. Neural networks are then used to produce rendering parameters such as colour, opacity, geometry, or approximate local mapping functions such as deformation and motion. The approach improves real-time performance, memory efficiency, and fidelity, but leaves open challenges in initialization, balancing workloads, maintaining high-frequency detail, scaling to large scenes, and enabling intuitive editing.

\subsection{Applications of NeRF and Neural Field Rendering post-Gaussian Splatting}
With the rapid development of Large Language Models, Vision Language Models \cite{2021clipvit} and pretrained 2D foundational models \cite{SAM, dino}, 3D scene understanding and 3D grounding have emerged as a new area of research. However, despite some recent advances in this area, the field of grounded 3D scene representation and Vision Language Model-based 3D representation research, including 3D question answering and semantic understanding, is largely dominated by Gaussian Splatting-based methods \cite{2024feisurveyFeb}. 

SLAM and 3D human avatars remained a popular area of research. One possible reason is that implicit and hybrid neural field representations can be advantageous for these two applications. These representations require less memory and storage and are easier to query (by simply calling the neural field at a particular point in space), as opposed to a point-cloud-like 3D Gaussian Splatting representation. 3D fields can also arise naturally from the framework's formulation, as is the case with articulated human avatar modeling.

In SLAM, the implicit and hybrid representations have lower memory and storage requirements than explicit representations, such as Gaussian Splatting. This can be relevant for methods designed to perform SLAM onboard the platform itself. Additionally, as presented in Section \ref{sub:4}, an emerging area of research is combining SLAM with autonomous agent-based navigation (as opposed to user-controlled navigation). Certain autonomous navigation algorithms may prefer easy-to-query implicit 3D representations to a 3D point-cloud-like Gaussian Splatting representation \cite{yang2024slam}.

For 3D human avatars, the dominant paradigm is to build neural fields on top of a baseline articulated SMPL \cite{2015SMPL} skeleton model. NeRF and adjacent methods fit this framework more naturally. Many Gaussian Splatting methods built upon SMPL use a combination of neural fields and Gaussian primitive representations \cite{2024hugs, 2024human, 2024gauhuman}. Therefore, it is not surprising that implicit and hybrid neural field methods remain popular in this field of research.

In terms of computer graphics and computer vision research, 3D reconstruction with neural implicit signed distance functions is also an area of work that stays active post-Gaussian Splatting. Signed distance functions support meshing pipelines such as TSDF and NeRF frameworks that rely on implicit signed distance fields, for example \cite{2021neus}, which can yield meshes of higher fidelity than Gaussian Splatting-based meshing as shown in \cite{2023neuralangelo}. NeRF is well adapted to model local continuous field functions with a compact neural network and this includes lighting fields, deformation fields with dynamic structure, and signed distance fields for continuous surface modeling. NeRF also remains strong for physically based inverse rendering that recovers real scene properties such as BRDF and lighting, as demonstrated in \cite{2023renerf}. Its volume rendering formulation gives a direct route to occlusion and light transport, and this supports relighting and material editing with near first principles behavior.

\section{Conclusion} \label{sec:conclusion}
Since the original paper by Mildenhall et al., NeRF and implicit/hybrid neural field rendering methods have made tremendous progress in terms of speed, quality, and training view requirements, addressing the weaknesses of the original model. NeRF models have found numerous applications in areas such as urban mapping, photogrammetry, image editing, labeling, processing, and 3D reconstruction and view synthesis of human avatars and urban environments. Although the research interest of the computer vision community has shifted towards Gaussian Splatting in many key research areas, there remains much interest in NeRF and implicit/hybrid neural field rendering in applications where implicit/hybrid representation, or a volumetric rendering approach, is advantageous. Moreover, many Gaussian Splatting methods drew inspiration from earlier NeRF methods. By studying earlier NeRF and neural field rendering papers, future authors may find further inspiration for other novel view synthesis-based research.

NeRF is an exciting paradigm for novel view synthesis, 3D reconstruction, 3D scene representation, and applications thereof. By providing this survey, we aim to introduce more computer vision practitioners to this field, provide a helpful reference of existing NeRF models and datasets, and motivate future research with our discussions.